%% file: main.tex
\documentclass[10pt,twocolumn,letterpaper]{article}

\usepackage{cvpr}              % To produce the CAMERA-READY version
\input{preamble}

\definecolor{cvprblue}{rgb}{0.21,0.49,0.74}
\usepackage[pagebackref,breaklinks,colorlinks,allcolors=cvprblue]{hyperref}

\def\paperID{*****} % *** Enter the Paper ID here
\def\confName{CVPR}
\def\confYear{2026}

\begin{document}
\title{The Eleventh NTIRE 2026 Efficient Super-Resolution Challenge Report}

%%%%%%%%% AUTHORS - PLEASE UPDATE
\input{sec/authors}

\input{sec/0_abstract}    
\input{sec/1_intro}
\input{sec/2_ntire26_challenge}
\input{sec/3_results}

\section{Challenge Methods and Teams}
\label{sec:methods_and_teams}
\input{teams/team22_XiaomiMM/main}
\input{teams/team01_BOE_AIoT/main}
\input{teams/team16_PKDSR/main}

\input{teams/team04_ZenoSR/main}
\input{teams/team05_CUIT_HTT/main}

\input{teams/team06_HAESR/main}
\input{teams/team09_IN2GM/main}
\input{teams/team10_Sunflower/main}
\input{teams/team11_XuptSR/main}
\input{teams/team12_WMESR/main}
\input{teams/team15_VARH-AI/main}

\input{teams/team17_XSR/main}
\input{teams/team18_DISP/main}
\input{teams/team20_JustTry/main}

\input{teams/team21_MDAP/main}

% \clearpage
\section*{Acknowledgments}
This work was partially supported by the Humboldt Foundation. We thank the NTIRE 2025 sponsors: ByteDance, Meituan, Kuaishou, and University of Wurzburg (Computer Vision Lab).

\input{sec/5_appendix}

{
\small
\bibliographystyle{unsrt}
\bibliography{main}
}


\end{document}

%% file: preamble.tex
\usepackage{colortbl}%
\usepackage{graphicx}
\usepackage{amsmath}
\usepackage{amssymb}
\usepackage{booktabs,caption}
\usepackage{multirow}   
\usepackage{bm}
\usepackage{dsfont}
\usepackage[linesnumbered,ruled,vlined]{algorithm2e}
\usepackage{makecell}
\usepackage[flushleft]{threeparttable}
\usepackage{enumitem}
\usepackage{siunitx}
\usepackage[accsupp]{axessibility} % Improves PDF readability for those with disabilities.

\definecolor{MyBlack}{HTML}{323A45}

\def\eg{\emph{e.g.,\ }}
\def\ie{\emph{i.e.,\ }}

%% file: sec/authors.tex
%-----------------------%
% Author Information & Maketitle
%-----------------------%
\author{
Bin Ren$^*$ \and
Hang Guo$^*$ \and
Yan Shu$^*$ \and
Jiaqi Ma$^*$ \and
Ziteng Cui$^*$ \and
Shuhong Liu$^*$ \and 
Guofeng Mei$^*$ \and
Lei Sun$^*$ \and 
Zongwei Wu$^*$ \and 
Fahad Shahbaz Khan$^*$ \and 
Salman Khan$^*$ \and 
Radu Timofte$^*$ \and 
Yawei Li$^*$ \and
Hongyuan Yu \and
Pufan Xu  \and
Chen Wu \and
Long Peng \and
Jiaojiao Yi \and
Siyang Yi \and
Yuning Cui \and
Jingyuan Xia \and
Xing Mou \and
Keji He \and
Jinlin Wu \and
Zongang Gao \and
Sen Yang \and
Rui Zheng \and
Fengguo Li \and
Yecheng Lei \and
Wenkai Min \and
Jie Liu \and 
Keye Cao \and 
Shubham Sharma \and
Manish Prasad \and
Haobo Li \and
Matin Fazel \and
Abdelhak Bentaleb \and
Rui Chen \and
Shurui Shi \and
Zitao Dai \and
Qingliang Liu \and
Yang Cheng \and
Jing Hu \and
Xuan Zhang \and
Rui Ding \and 
Tingyi Zhang \and
Hui Deng \and
Mengyang Wang \and 
Fulin Liu \and 
Jing Wei \and 
Qian Wang \and 
Hongying Liu \and 
Mingyang Li \and 
Guanglu Dong \and 
Zheng Yang \and 
Chao Ren \and 
Hongbo Fang \and 
Lingxuan Li \and 
Lin Si \and 
Pan Gao \and 
Moncef Gabbouj \and 
Watchara Ruangsang \and 
Supavadee Aramvith
}
\maketitle
\let\thefootnote\relax\footnotetext{
$^*$ Bin Ren, Hang Guo, Yan Shu, Jiaqi Ma, Ziteng Cui, Shuhong Liu, Guofeng Mei, Lei Sun, Zongwei Wu, Fahad Shahbaz Khan, Salman Khan, Radu Timofte, and Yawei Li were the challenge organizers, while the other authors participated in the challenge.\\ 
Appendix~\ref{sec:teams} contains the authors' teams and affiliations.\\
NTIRE 2026 webpage: \url{https://cvslai.net/ntire/2026/}.\\ 
Code: \url{https://github.com/Amazingren/NTIRE2026_ESR/}.
}

%% file: sec/0_abstract.tex
%%%%%%%%% ABSTRACT
\begin{abstract}
This paper reviews the NTIRE 2026 challenge on efficient single-image super-resolution with a focus on the proposed solutions and results. 
The aim of this challenge is to devise a network that reduces one or several aspects, such as runtime, parameters, and FLOPs, while maintaining PSNR of around 26.90 dB on the DIV2K\_LSDIR\_valid dataset, and 26.99 dB on the DIV2K\_LSDIR\_test dataset. 
The challenge had 95 registered participants, and 15 teams made valid submissions. They gauge the state-of-the-art results for efficient single-image super-resolution.
\end{abstract}  

%% file: sec/1_intro.tex
%%%%%%%%% BODY TEXT
\section{Introduction}
\label{sec:introduction}
Single-image super-resolution (SR) aims to recover a high-resolution (HR) image from a single low-resolution (LR) observation degraded by a known process.
In most SR settings, the LR image is assumed to be generated through a combination of blurring and down-sampling. Under the classical formulation, bicubic down-sampling is widely adopted as a standard degradation model, providing a consistent benchmark for evaluating and comparing different SR approaches.

Super-resolution has been widely applied in a range of practical scenarios, including mobile photography, video streaming, surveillance, medical imaging, and remote sensing~\cite{xie2025mat,zhang2025perceive,shu2026terrascope,ren2024sharing,tang2025degradation,miao2025langhops,xie2025star,georgescu2023multimodal,xiao2024frequency}. In many of these cases, acquiring high-resolution images is constrained by hardware, bandwidth, or storage limitations. SR methods therefore, offer a flexible alternative by improving image quality at the reconstruction stage.

Recent advances in deep neural networks have significantly improved SR performance. However, these improvements are often accompanied by increased model complexity, including a large number of parameters, high computational cost, and noticeable inference latency. Such characteristics limit their applicability, particularly on resource-constrained platforms where real-time processing is required. To address these challenges, a growing body of work has focused on improving model efficiency. Representative approaches include network pruning, low-rank decomposition, quantization, neural architecture search, state-space models such as Mamba, and knowledge distillation~\cite{vim,liu2018rethinking,zheng2026open,zheng2025distilling,ren2023masked,yu2017compressing,ma2024shapesplat,zhao2024denoising,an2025onestory,li2026chorus,lobba2025inverse,wang2025aim}. These techniques aim to reduce computational overhead while preserving reconstruction quality, and have been increasingly explored in the SR domain.

The efficiency of a deep neural network can be evaluated from multiple perspectives, including runtime, parameter count, computational complexity (FLOPs), memory usage, and activation size. Among these, runtime serves as the most direct indicator of practical efficiency. Computational complexity is closely related to energy consumption, which is critical for battery-powered devices, while memory footprint and parameter size directly affect deployment feasibility and hardware cost.

In conjunction with the NTIRE 2026 workshop on New Trends in Image Restoration and Enhancement, we organize the Efficient Super-Resolution challenge as part of the NTIRE ESR track. The goal is to reconstruct $\times 4$ HR images from LR inputs using models that achieve a better balance between performance and efficiency. Participants are expected to design methods that reduce runtime, parameter count, and FLOPs compared to the SPAN baseline~\cite{wan2024swift}, while maintaining at least 26.90 dB on the DIV2K\_LSDIR\_valid dataset and 26.99 dB on the DIV2K\_LSDIR\_test dataset.

This challenge aims to provide a standardized benchmark for efficient SR, encourage the development of practically deployable solutions, and offer insights into effective design strategies for efficient SR networks.

This challenge is one of the challenges associated with the NTIRE 2026 Workshop~\footnote{\url{https://www.cvlai.net/ntire/2026/}} on:
Deepfake detection~\cite{ntire26deepfake}, 
high-resolution depth~\cite{ntire26hrdepth},
multi-exposure image fusion~\cite{ntire26raim_fusion}, 
AI flash portrait~\cite{ntire26raim_portrait}, 
professional image quality assessment~\cite{ntire26raim_piqa},
light field super-resolution~\cite{ntire26lightsr},
3D content super-resolution~\cite{ntire263dsr},
bitstream-corrupted video restoration~\cite{ntire26videores},
X-AIGC quality assessment~\cite{ntire26XAIGCqa},
shadow removal~\cite{ntire26shadow},
ambient lighting normalization~\cite{ntire26lightnorm},
controllable Bokeh rendering~\cite{ntire26bokeh},
rip current detection and segmentation~\cite{ntire26ripdetseg},
low light image enhancement~\cite{ntire26llie},
high FPS video frame interpolation~\cite{ntire26highfps},
Night-time dehazing~\cite{ntire26nthaze,ntire26nthaze_rep},
learned ISP with unpaired data~\cite{ntire26isp},
short-form UGC video restoration~\cite{ntire26ugcvideo},
raindrop removal for dual-focused images~\cite{ntire26dual_focus},
image super-resolution (x4)~\cite{ntire26srx4},
photography retouching transfer~\cite{ntire26retouching},
mobile real-word super-resolution~\cite{ntire26rwsr},
remote sensing infrared super-resolution~\cite{ntire26rsirsr},
AI-Generated image detection~\cite{ntire26aigendet},
cross-domain few-shot object detection~\cite{ntire26cdfsod},
financial receipt restoration and reasoning~\cite{ntire26finrec},
real-world face restoration~\cite{ntire26faceres},
reflection removal~\cite{ntire26reflection},
anomaly detection of face enhancement~\cite{ntire26anomalydet},
video saliency prediction~\cite{ntire26videosal},
efficient super-resolution~\cite{ntire26effsr},
3d restoration and reconstruction in adverse conditions~\cite{ntire26realx3d},
image denoising~\cite{ntire26denoising},
blind computational aberration correction~\cite{ntire26aberration},
event-based image deblurring~\cite{ntire26eventblurr},
efficient burst HDR and restoration~\cite{ntire26bursthdr},
low-light enhancement: `twilight Cowboy'~\cite{ntire26twilight},
and efficient low light image enhancement~\cite{ntire26effllie}.

%% file: sec/2_ntire26_challenge.tex
\section{NTIRE 2026 Efficient Super-Resolution Challenge}
The goals of this challenge include: (1) promoting research in the area of efficient super-resolution, (2) facilitating comparisons between the efficiency of various methods, and (3) providing a platform for academic and industrial participants to engage, discuss, and potentially establish collaborations. This section delves into the specifics of the challenge.

\subsection{Dataset}
The DIV2K~\cite{agustsson2017ntire} dataset and LSDIR~\cite{lilsdir} dataset are utilized for this challenge. The DIV2K dataset consists of 1,000 diverse 2K resolution RGB images, which are split into a training set of 800 images, a validation set of 100 images, and a test set of 100 images. The LSDIR dataset contains 86,991 high-resolution high-quality images, which are split into a training set of 84,991 images, a validation set of 1,000 images, and a test set of 1,000 images. In this challenge, the corresponding LR DIV2K images are generated by bicubic downsampling with a down-scaling factor of 4$\times$. 
The training images from DIV2K and LSDIR are provided to the participants of the challenge. During the validation phase, 100 images from the DIV2K validation set and 100 images from the LSDIR validation set are made available to participants. 
During the test phase, 100 images from the DIV2K test set and another 100 images from the LSDIR test set are used. 
Throughout the entire challenge, the testing HR images remain hidden from the participants.

\subsection{SPAN Baseline Model}
\label{sec:baseline_model}

The Swift Parameter-free Attention Network  (SPAN)~\cite{wan2024swift} serves as the baseline model in this challenge. 
The aim is to improve its efficiency in terms of runtime, number of parameters, and FLOPs, while at least maintaining 26.90 dB on the DIV2K\_LSDIR\_valid dataset and 26.99 dB on the DIV2K\_LSDIR\_test dataset.

The main idea within SPAN is that a parameter-free attention mechanism is constructed by passing extracted features through a symmetric activation function around the origin to calculate attention maps directly. This attention mechanism focuses on
information-rich regions without the need of additional parameter learning, allowing for rapid and effective feature extraction from shallow to deep layers. The design of symmetric activation functions and residual connections in the modules helps to solve issues related to information loss of the parameter-free attention modules. The simplicity of the network structure ensures operational speed, addressing the
challenges posed by conventional attention mechanisms.

The baseline SPAN emerges as the 1st place for the overall performance of the NTIRE2024 Efficient SR Challenge~\cite{wan2024swift}. The quantitative performance and efficiency metrics of SPAN are summarized as follows:
(1) The number of parameters is 0.151 M. 
(2) The average PSNRs on validation (DIV2K 100 valid images and LSDIR 100 valid images) and testing (DIV2K 100 test images and LSDIR 100 test images) sets of this challenge are 26.94 dB and 27.01 dB, respectively. 
(3) The runtime averaged to 7.65ms on the validation and test set with PyTorch 2.5.1+cu124, and a single NVIDIA RTX A6000 GPU. 
(4) The number of FLOPs for an input of size $256\times256$ is 9.83 G.

\subsection{Tracks and Competition}
The aim of this challenge is to devise a network that reduces one or several aspects such as runtime, parameters, and FLOPs, while at least maintaining the 26.90 dB on the DIV2K\_LSDIR valid dataset, and 26.99 dB on the DIV2K\_LSDIR test dataset. 

\medskip
\noindent{\textbf{Challenge phases: }}
\textit{(1) Development and validation phase}: Participants were given access to 800 LR/HR training image pairs and 200 LR/HR validation image pairs from the DIV2K and the LSDIR datasets. 
An additional 84,991 LR/HR training image pairs from the LSDIR dataset are also provided to the participants. 
The SPAN model, pre-trained parameters, and validation demo script are available on GitHub \url{https://github.com/Amazingren/NTIRE2026_ESR}, allowing participants to benchmark their models' runtime on their systems. 
Participants could upload their HR validation results to the evaluation server to calculate the PSNR of the super-resolved image produced by their models and receive immediate feedback. 
The corresponding number of parameters, FLOPs, and runtime will be computed by the participants.
\textit{(2) Testing phase}: In the final test phase, participants were granted access to 100 LR testing images from DIV2K and 100 LR testing images from LSDIR, while the HR ground-truth images remained hidden. Participants submitted their super-resolved results to the Codalab evaluation server and emailed the code and factsheet to the organizers. 
The organizers verified and ran the provided code to obtain the final results, which were then shared with participants at the end of the challenge.

\medskip
\noindent{\textbf{Evaluation protocol: }}
Quantitative evaluation metrics included validation and testing PSNRs, runtime, number of parameters, and FLOPs during inference. 
PSNR was measured by discarding a 4-pixel boundary around the images. The average runtime during inference was computed on the 200 LR validation images and the 200 LR testing images.
The average runtime on the validation and testing sets served as the final runtime indicator. 
FLOPs are evaluated on an input image of size 256 $\times$ 256. 
Among these metrics, runtime was considered the most important. Participants were required to maintain a PSNR of at least 26.90 dB on the DIV2K\_LSDIR valid dataset, and 26.99 dB on the DIV2K\_LSDIR test dataset during the challenge.  
The constraint on the testing set helped prevent overfitting on the validation set. 
A code example for calculating these metrics is available at \url{https://github.com/Amazingren/NTIRE2026_ESR}. 
The codes and the corresponding pre-trained weights will be available in this repository later.

To better quantify the rankings, we followed the scoring function from previous NTIRE ESR~\cite{ren2024ninth,ren2025tenth} for three evaluation metrics in this challenge: 
runtime, 
FLOPs, and 
parameters. 
This scoring aims to convert the performance of each metric into corresponding scores to make the rankings more significant. Especially, the score for each separate metric (\ie Runtime, FLOPs, and parameter) for each sub-track is calculated as:
\begin{equation}
    \begin{aligned}
    Score\_{Metric} = \dfrac{\text{Exp} (2 \times Metric_{TeamX})}{Metric_{Baseline}},
    \end{aligned}
    \label{equ:score1}
\end{equation}
based on the score of each metric, the final score used for the main track is calculated as:
\begin{equation}
    \begin{aligned}
    Score\_{Final} & = w_{1} \times Score\_Runtime \\
    & + w_{2} \times Score\_FLOPs \\
    & + w_{3} \times Score\_Params,
    \end{aligned}
    \label{equ:score2}
\end{equation}
where $w_{1}$, $w_{2}$, and $w_{3}$ are set to 0.8, 0.1, and 0.1, respectively. This setting is intended to incentivize participants to design a method that prioritizes speed efficiency while maintaining a reasonable model complexity.

%% file: sec/3_results.tex
\section{Challenge Results}
\input{sec/results_table}
The final challenge results and the corresponding rankings are presented in Tab.~\ref{tbl:final_results}.
The table also includes the baseline method SPAN~\cite{wan2024swift} for comparison. 
In Sec.\ref{sec:methods_and_teams}, the methods evaluated in Tab.~\ref{tbl:final_results} are briefly explained, while the team members are listed in~\ref{sec:teams}. 
The performance of different methods is compared from four different perspectives, including 
the runtime, 
FLOPs, 
the parameters, 
and the overall performance.
Furthermore, in order to promote a fair competition emphasizing efficiency, the criteria for image reconstruction quality in terms of test PSNR are set to 26.90 and 26.99 on the DIV2K\_LSDIR\_valid and DIV2K\_LSDIR\_test sets, respectively.

\noindent\textbf{Runtime.} In this challenge, runtime stands as the paramount evaluation metric. \textbf{XiaomiMM}'s solution emerges as the frontrunner with the shortest runtime among all entries in the efficient SR challenge, securing its top-1 ranking position. Following closely, DISP and BOE\_AIoT claim the second and third spots, respectively. Remarkably, the average runtime of the top three solutions on both the validation and test sets remains below 7 ms. Impressively, the first 6 teams present solutions with an average runtime below 8 ms, showcasing a continuous enhancement in the efficiency of image SR networks. Despite the slight differences in runtime among the top three teams, the challenge retains its competitive edge. A notable characteristic of this year's challenge is that runtime plays a more decisive role in the overall rankings compared to previous years, as the weight of runtime in the scoring formula has been increased to encourage participants to design more efficient solutions such as quantization and pruning algorithms. This shift indicates that participants are now placing greater emphasis on runtime optimization while still maintaining competitive image reconstruction quality.

\noindent\textbf{Parameters.} Model complexity was further evaluated by considering the number of parameters, as detailed in Table~\ref{tbl:final_results}. In this sub-track, \textbf{ZenoSR} achieved the top position with only 0.038M parameters, closely followed by XuptSR and XSR with 0.051M and 0.086M parameters, respectively. The minimal disparity among the top three methods highlights their competitive edge and efficiency in managing model complexity. They were scored at 1.65, 1.97, and 3.12, respectively, indicating a tight competition. However, it is noteworthy that these models also exhibited relatively high runtimes, suggesting an area for potential improvement in future iterations.

\noindent\textbf{FLOPs.} The number of floating-point operations (FLOPs) is another critical metric for assessing model complexity. Within this sub-track, \textbf{ZenoSR}, XuptSR, and XSR secured the top three positions with FLOPs of 2.68G, 3.30G, and 5.24G, respectively. The competitiveness of this sub-track is further confirmed by the close scores of 1.73, 1.96, and 2.90, aligned with the parameter evaluation results. Remarkably, the same models top both the parameters and FLOPs evaluations, demonstrating consistent performance across different complexity metrics. Similar to the parameters sub-track, the extended runtimes of these methods point to a need for further research and optimization. Key implications include:
i) \textit{Efficiency vs. Performance Trade-off}: The close competition among the top models in terms of parameters and FLOPs suggests a significant trade-off between model efficiency and performance. Despite achieving minimal parameter counts and FLOPs, the high runtimes indicate that these models might be optimizing computational complexity at the expense of execution speed. This raises important considerations for future research in balancing efficiency with real-world usability, especially in applications where inference speed is critical.
ii) \textit{Potential for Model Optimization}: The consistency in ranking between the parameters and FLOPs sub-tracks reveals that models which are optimized for one aspect of computational efficiency tend to perform well in others. However, the noted high runtimes across these models suggest an untapped potential for holistic model optimization. Future work could focus on integrating more advanced optimization techniques or exploring novel architectural innovations to enhance both the computational efficiency and runtime performance.

\noindent\textbf{PSNR.} Team \textbf{HAESR}, WMESR, Sunflower, and XuptSR demonstrate superior PSNR values, a critical evaluation metric in super-resolution. Specifically, HAESR leads with an exceptional 27.22 dB,  followed by WMESR with 27.06 dB, and Sunflower and XuptSR both at 27.03 dB on the DIV2K\_LSDIR\_test set. Despite these impressive performances, it is essential to emphasize that the primary focus of this challenge is on \textit{efficiency in super-resolution}. Accordingly, we have adjusted the PSNR criteria, setting rigorous lower thresholds of 26.90 dB and 26.99 dB for the DIV2K\_LSDIR\_valid and DIV2K\_LSDIR\_test sets, respectively. This adjustment is designed to prioritize a balance between high performance and computational efficiency.
A commendable total of 14 teams met this adjusted benchmark, demonstrating their capability to effectively balance image quality with efficiency. However, team MDAP, while notable for their efficiency, did not achieve the required PSNR levels. This highlights the ongoing challenge of optimizing super-resolution processes that meet both efficiency and performance standards, underscoring the complex nature of advancements in this field.

\noindent\textbf{Overall Evaluation.} The final assessment of performance employs a comprehensive metric that synthesizes runtime, FLOPs, and the number of parameters into a unified score. In this rigorous evaluation, the \textbf{XiaomiMM} Group excelled, claiming the prestigious top position, followed by BOE\_AIoT and PKDSR in second and third places, respectively. This achievement highlights the sophisticated engineering and innovative approaches implemented by these groups.

Contrasting with the previous year, where a balanced approach across metrics influenced overall rankings, this year presents a notable shift back toward runtime dominance. The best performer in runtime also secured first place in the overall competition. Specifically, XiaomiMM, the overall winner, ranked first in runtime, fifth in parameters, and sixth in FLOPs. Similarly, BOE\_AIoT, which came second overall, was third in runtime, sixth in parameters, and seventh in FLOPs. This demonstrates that:
i) While a balanced approach to model design remains important, runtime efficiency continues to play a decisive role in overall rankings, particularly when top competitors achieve similarly competitive parameter and FLOPs counts.
ii) Achieving top performance in runtime, when combined with reasonably competitive complexity metrics, can effectively translate into strong overall rankings, underscoring the continued importance of inference speed in practical super-resolution applications.
This year's goal was to encourage a balanced pursuit of speed and efficiency, a challenge that has evidently led to significant innovations and advancements in model design.

\subsection{Main Ideas}
Throughout this challenge, several techniques have been proposed to enhance the efficiency of deep neural networks for image super-resolution (SR) while striving to maintain optimal performance. The choice of techniques largely depends on the specific metrics that a team aims to optimize. Below, we outline some typical ideas that have emerged:

\begin{itemize}
    \item \textbf{Hardware-level optimization through custom CUDA kernel fusion emerges as a novel and impactful direction.} XiaomiMM, the overall winner, demonstrated that for small efficient SR networks, the performance bottleneck lies not in FLOPs but in memory bandwidth. Their proposed \texttt{span\_attn\_op} fuses the 1$\times$1 attention convolution, element-wise addition, and element-wise multiplication into a single CUDA kernel, eliminating 3$\times$ redundant DRAM round-trips. This low-level optimization, combined with architectural improvements in SPANV2, contributed directly to achieving the shortest runtime among all submissions, underscoring the importance of hardware-aware design in practical deployment scenarios.
    
    \item \textbf{Network pruning combined with knowledge distillation remains a practical and effective strategy for model compression.} Several teams adopted channel pruning on strong baseline models to reduce runtime and parameter count, relying on distillation to recover the lost reconstruction quality. BOE\_AIoT pruned the final layer of SPANF from 32 to 20 channels and employed knowledge distillation for fine-tuning. Similarly, PKDSR performed a gradual two-stage pruning with separate distillation rounds at each stage, demonstrating that progressive pruning leads to more stable training and smaller performance degradation.
    
    \item \textbf{Knowledge distillation, as an independent training strategy beyond pruning, is widely explored.} Beyond its use alongside pruning, distillation is applied as a standalone technique for lightweight model training. VARH-AI employed DSCLoRA distillation with spatial affinity alignment, transferring knowledge through Separable Convolutional LoRA Blocks. CUIT\_HTT utilized a SwinIR teacher to supervise a compact SPAN-style student via hybrid pixel and edge distillation losses. WMESR adopted a progressive 4-stage knowledge distillation pipeline (pre-training, large-patch fine-tuning, PSNR-oriented fine-tuning, and distillation) to bridge the quality gap of their aggressively lightweight DWMamba. DISP also applied affinity-based distillation between teacher and student networks.

    \item \textbf{Re-parameterization is commonly adopted to improve inference speed without sacrificing model capacity during training.} Several teams incorporated structural re-parameterization, where multi-branch training-time architectures (\eg, ERB and MBRB in ERRN2 by Just Try, or reparameterized convolutions in DISP) are collapsed into a single 3$\times$3 convolution at inference. VARH-AI further extended this idea via operator fusion, mathematically compacting sequential 3$\times$3 convolutions into equivalent 5×5 kernels, eliminating redundant parameters at zero inference cost.

    \item \textbf{State space models (Mamba) continue to be explored as a complement to CNN-based SR architectures.} CUIT HTT proposed MambaGate-SR, which preserves an efficient CNN backbone while introducing a lightweight Mamba-based gating branch for global context modulation, effectively combining local convolutional modeling with global dependency capturing. WMESR's DWMamba adopted a frequency-decoupled design, using Discrete Wavelet Transform (DWT) to decompose features into low- and high-frequency subbands, with a Mamba-based branch processing the low-frequency subband and a lightweight CNN branch handling the high-frequency components, followed by Cross-Frequency Interaction for structural guidance.
    
    \item \textbf{Various other techniques are also attempted.} Several teams explored attention-sharing mechanisms to reduce redundant computation: HAESR proposed Shared Hybrid Attention Blocks that reuse attention maps generated by a preceding Hybrid Attention Block, combining Meso-scale and Global spatial attention. The parameter-free attention mechanism from SPAN continues to inspire new designs: IN2GM replaced the enhanced spatial attention module in RFDN with a parameter-free attention module to reduce parameters and FLOPs, while XiaomiMM redesigned it into a learned 1$\times$1 projection for richer channel mixing. Advanced training strategies such as multi-stage progressive training, FFT-based frequency loss, and EMA weight averaging were also widely adopted across teams to further boost performance.
\end{itemize}

\subsection{Fairness}
To ensure the fairness of the efficient SR challenge, several rules were established, mainly concerning the dataset used for training the network. First, training with additional external datasets, such as Flickr2K, was allowed. Second, training with the additional DIV2K validation set, including either HR or LR images, was not permitted, as the validation set was used to assess the overall performance and generalizability of the proposed network. 
Third, training with DIV2K test LR images was prohibited. 
Lastly, using advanced data augmentation strategies during training was considered a fair approach.

\subsection{Conclusions}
The analysis of the submissions to this year's Efficient SR Challenge allows us to draw several important conclusions:

\begin{itemize}
    \item Firstly, the SR community continues to demonstrate strong research vitality in the area of efficient super-resolution. This year, the challenge attracted \textbf{97} registered participants, with \textbf{15 }teams making valid submissions. Despite a smaller number of participants compared to previous years, all proposed methods have meaningfully advanced the state-of-the-art in efficient SR, reflecting a maturing and more focused research community.
    \item Secondly, in contrast to last year's trend where balanced solutions across all metrics proved most beneficial, runtime efficiency has re-emerged as the decisive factor in overall rankings this year. The increased weight of runtime in the scoring formula ($w_1$ = 0.8) incentivized participants to prioritize inference speed. Accordingly, the team achieving the shortest runtime (XiaomiMM) also claimed first place overall, demonstrating that under this evaluation protocol, excelling in runtime while maintaining reasonably competitive parameter counts and FLOPs translates most effectively into top rankings.
    \item Thirdly, network pruning combined with knowledge distillation has proven to be a reliable and practical compression strategy. Teams such as BOE\_AIoT and PKDSR demonstrated that carefully pruning channel dimensions of a strong baseline model, guided by distillation from the unpruned teacher, can yield competitive runtime and parameter reductions with minimal PSNR degradation. This combination remains one of the most accessible yet effective approaches for efficient SR.
    \item Fourthly, hardware-aware optimization has emerged as a new and impactful frontier beyond architectural design. The winning solution SPANV2 from XiaomiMM introduced a custom fused CUDA kernel that eliminates redundant DRAM memory round-trips in the attention computation, achieving the best runtime without relying solely on reducing FLOPs or parameters. This result highlights that for highly optimized lightweight networks, the performance bottleneck increasingly lies in memory bandwidth rather than arithmetic complexity, and that low-level hardware-software co-design is becoming an essential dimension of efficient SR research.
    \item Fifthly, consistent with previous challenges, the use of large-scale datasets such as LSDIR~\cite{lilsdir} for training, combined with multi-stage progressive training strategies, continues to play an important role in achieving high reconstruction quality. Most teams adopted multi-phase training pipelines with progressively increasing patch sizes and decreasing learning rates, which has become a standard practice in the community for optimizing both convergence stability and final PSNR performance.
    \item Sixthly, while the state space model (Mamba) continues to attract exploration as a promising architectural component, its practical advantage in runtime-oriented challenges remains limited. Two teams (CUIT\_HTT and WMESR) incorporated Mamba-based modules into their SR frameworks, validating its capability for global dependency modeling. However, both solutions exhibited relatively high runtimes compared to the top-ranked entries, suggesting that further research is needed to unlock the runtime potential of Mamba-based designs in efficient SR.
    \item Seventhly, a notable efficiency paradox is observed between parameter/FLOPs efficiency and actual runtime. Teams such as ZenoSR and XuptSR achieved the lowest parameter counts and FLOPs in their respective sub-tracks, yet ranked near the bottom in runtime. This underscores that minimizing parameters and FLOPs does not necessarily translate into faster inference, and that runtime including operator efficiency, memory access patterns, and hardware utilization—requires dedicated attention beyond simply reducing arithmetic complexity.
\end{itemize}

Overall, as computational hardware continues to advance, the synergy between algorithmic innovation and hardware-aware design will become ever more critical. The introduction of custom CUDA kernels, novel distillation pipelines, and frequency-domain processing in this year's challenge reflects the broadening toolkit available to the efficient SR community. We expect continued progress driven by both architectural creativity and system-level optimization, with runtime efficiency remaining the central benchmark for practical deployment of super-resolution models.

%% file: sec/results_table.tex
\begin{table*}[!ht]
% \scriptsize
\centering
\setlength{\extrarowheight}{0.7pt}
\setlength{\tabcolsep}{16pt}
\caption{Results of Ninth NTIRE 2026 Efficient SR Challenge. The performance of the solutions is compared thoroughly from three perspectives including the runtime, FLOPs, and the number of parameters. The underscript numbers associated with each metric score denote the ranking of the solution in terms of that metric. For runtime, ``Val.'' is the runtime averaged on DIV2K\_LSDIR\_valid validation set. ``Test'' is the runtime averaged on a test set with 200 images from DIV2K\_LSDIR\_test set, respectively. ``Ave.'' is averaged on the validation and test datasets. 
``\#Params'' is the total number of parameters of a model. ``FLOPs'' denotes the floating point operations. 
Main Track combines all three evaluation metrics.
The ranking for the main track is based on the score calculated via Eq.~\ref{equ:score2}, and the ranking for other sub-tracks is based on the score of each metric via Eq.~\ref{equ:score1}. Please note that \textbf{this is not a challenge for PSNR improvement. The ``validation/testing PSNR'' is not ranked. For all the scores, the lower, the better.}
}
\label{tbl:final_results}
\begin{threeparttable}
    \resizebox{\linewidth}{!}
    {
    \begin{tabular}{@{\extracolsep{\fill}} 
                                    l|
                                    |
                                    S[table-format=2.2]
                                    S[table-format=2.2]
                                    |
                                    S[table-format=3.3] 
                                    S[table-format=3.3] 
                                    S[table-format=3.3]
                                    |
                                    S[table-format=1.3] 
                                    S[table-format=2.2] 
                                    |
                                    S[table-format=3.2$_{(3)}$] 
                                    S[table-format=3.2$_{(3)}$] 
                                    S[table-format=2.2$_{(3)}$]
                                    |
                                    S[table-format=3.2$_{(3)}$]
                                    S[table-format=2]}
                                    
    \toprule[1.0pt]
    \multirow{2}{*}{Teams}      & \multicolumn{2}{c|}{PSNR [dB]} & \multicolumn{3}{c|}{Runtime [ms]}  & {\#Params}  &   {FLOPs}  & \multicolumn{3}{c|}{Sub-Track Scores}  &\multicolumn{2}{c}{Main-Track}   \\ \cline{2-6} \cline{9-13}
    &{Val.} & {Test} & {Val.} & {Test} & {Ave.}  & {[M]} & {[G]} & {Runtime} & {\#Params} & {FLOPs} & {Overall Score} & {Ranking}  \\ 
    \midrule[0.6pt]
    XiaomiMM & 26.92 & 27.00 & 5.700 & 4.810 & 5.256 & 0.139 & 9.11 & 3.95$_{(1)}$ & 6.30$_{(5)}$ & 6.38$_{(6)}$ & 4.43 & \textbf{1} \\
    BOE\_AIoT & 26.90 & 26.99 & 6.942 & 6.500 & 6.724 & 0.142 & 9.25 & 5.80$_{(3)}$ & 6.56$_{(6)}$ & 6.57$_{(7)}$ & 5.95 & \textbf{2}\\
    PKDSR & 26.91 & 27.00 & 6.946 & 6.568 & 6.758 & 0.144 & 9.40 & 5.85$_{(4)}$ & 6.73$_{(7)}$ & 6.77$_{(8)}$ & 6.03 & \textbf{3} \\
    DISP & 26.90 & 27.02 & 6.650 & 6.238 & 6.444 & 0.164 & 10.69 & 5.39$_{(2)}$ & 8.78$_{(9)}$ & 8.80$_{(9)}$ & 6.07 & 4 \\
    VARH-AI & 26.92 & 26.99 & 8.472 & 7.314 & 7.892 & 0.126 & 8.22 & 7.87$_{(5)}$ & 5.31$_{(4)}$ & 5.33$_{(4)}$ & 7.36 & 5 \\
    Just Try & 26.90 & 27.01 & 9.886 & 8.892 & 9.388 & 0.246 & 16.06 & 11.64$_{(6)}$ & 26.00$_{(10)}$ & 26.25$_{(10)}$ & 14.54 & 6 \\
    IN2GM & 26.91 & 26.98 & 17.332 & 15.942 & 16.640 & 0.346 & 22.60 & 77.50$_{(7)}$ & 97.79$_{(11)}$ & 99.30$_{(12)}$ & 81.71 & 7 \\
    XSR & 26.90 & 27.02 & 18.886 & 17.808 & 18.348 & 0.086 & 5.24 & 121.13$_{(8)}$ & 3.12$_{(3)}$ & 2.90$_{(3)}$ & 97.51 & 8 \\
    Sunflower & 26.92 & 27.03 & 34.862 & 32.604 & 33.734 & 0.144 & 8.91 & \num{6.76e3}$_{(9)}$ & 6.73$_{(7)}$ & 6.13$_{(5)}$ & \num{5.41e3} & 9 \\
    ZenoSR & 26.90 & 27.01 & 50.108 & 46.184 & 48.144 & 0.038 & 2.68 & \num{2.93e5}$_{(11)}$ & 1.65$_{(1)}$ & 1.73$_{(1)}$ & \num{2.34e5} & 10 \\
    CUIT\_HTT & 26.90 & 27.00 & 39.778 & 37.560 & 38.670 & 1.106 & 33.40 & \num{2.46e4}$_{(10)}$ & \num{2.30e6}$_{(14)}$ & 893.84$_{(14)}$ & \num{2.50e5} & 11 \\
    XuptSR & 26.90 & 27.03 & 53.414 & 49.574 & 51.492 & 0.051 & 3.30 & \num{7.02e5}$_{(12)}$ & 1.97$_{(2)}$ & 1.96$_{(2)}$ & \num{5.62e5} & 12 \\
    HAESR & 27.08 & 27.22 & 88.208 & 84.058 & 86.132 & 0.353 & 26.61 & \num{6.02e9}$_{(13)}$ & 107.29$_{(12)}$ & 224.54$_{(13)}$ & \num{4.82e9} & 13 \\
    WMESR & 26.94 & 27.06 & 111.404 & 106.088 & 108.744 & 0.674 & 19.34 & \num{2.22e12}$_{(14)}$ & \num{7.53e3}$_{(13)}$ & 51.16$_{(11)}$ & \num{1.78e12} & 14 \\
    \midrule[0.6pt]
    \multicolumn{13}{c}{The following methods are not ranked since their validation/testing PSNR (underlined) is not on par with the threshold.}\\ 
    \midrule[0.6pt]
    MDAP & 26.80 & 26.91 & 131.948 & 127.458 & 129.702 & 0.149 & 9.62 & \num{5.33e14} & 7.20E+00 & 7.08E+00 & \num{4.26e14} & {-} \\
    EfficientSR Organizers & 26.94 & 27.01 & 7.858 & 7.436 & 7.650 & 0.151 & 9.83 & 7.39 & 7.39 & 7.39 & 7.39 & {-}\\
    \bottomrule[1pt]
    \end{tabular}
    }
\end{threeparttable}
\end{table*}

%% file: teams/team22_XiaomiMM/main.tex
\subsection{XiaomiMM}
\begin{figure*}[!t]
  \centering
  \includegraphics[width=\textwidth]{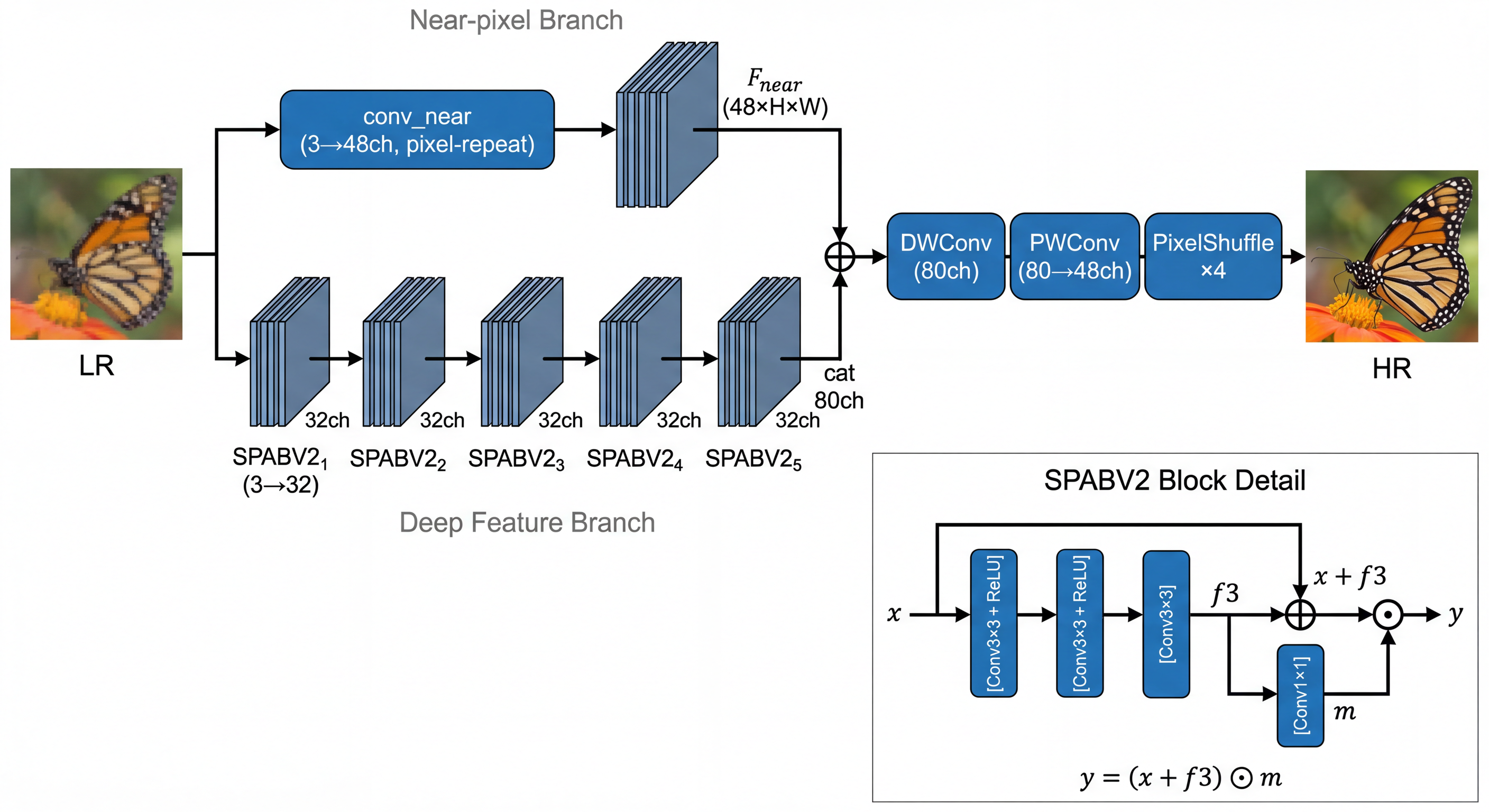}
  \caption{Team \textit{XiaomiMM}: SPANV2 overall architecture. The near-pixel branch (top)
    provides a pixel-repeat upsampling prior, while five SPABV2 blocks (bottom)
    extract deep features. The two paths are concatenated (80\,ch) and
    fused by depthwise-separable convolution before PixelShuffle$\times4$.}
  \label{fig:spanv2_arch}
\end{figure*}

\textbf{Method Description.} XiaomiMM presents \textbf{SPANV2} (\emph{Swift Parameter-free Attention Network
V2}), its submission to the NTIRE~2026 Efficient SR Challenge, built on
SPAN~\cite{wan2024swift} and SPANF~\cite{ren2025tenth}.

SPANV2 introduces three targeted contributions.
\textbf{(1) Learned attention.}
SPAN's parameter-free attention computes $\mathbf{y}=\mathbf{f}_1\!\odot\!\mathbf{f}_3$:
the weights are non-negative and restricted to element-wise products of two
fixed intermediate features, providing no dedicated channel-mixing capacity.
SPANV2 redesigns the core block into \emph{SPABV2}, replacing this mechanism with a
learned $1\!\times\!1$ projection that maps $\mathbf{f}_3$ to a full
$C\!\times\!C$ channel-mixing map $\mathbf{m}$, enabling content-adaptive
suppression and cross-channel gating with only $C^2$ additional parameters per block.
\textbf{(2) Fused CUDA kernel.}
For small efficient SR networks, the bottleneck lies not in FLOPs but in memory
bandwidth: each of the three operations in the attention step reads and writes
the full $C\!\times\!H\!\times\!W$ feature tensor to DRAM, resulting in
very low arithmetic intensity.
\texttt{span\_attn\_op} fuses all three operations into a single CUDA kernel,
eliminating $3\!\times$ redundant DRAM round-trips.
\textbf{(3) Near-pixel upsampling branch.}
Because low-frequency content dominates natural images, SPANV2 incorporates this prior
by initializing a parallel depthwise branch as exact nearest-neighbor
upsampling (equivalent to a $1\!\times\!1$ conv at initialization), allowing
the deep branch to focus on high-frequency residuals.

\begin{figure}[!t]
  \centering
  \includegraphics[width=\columnwidth]{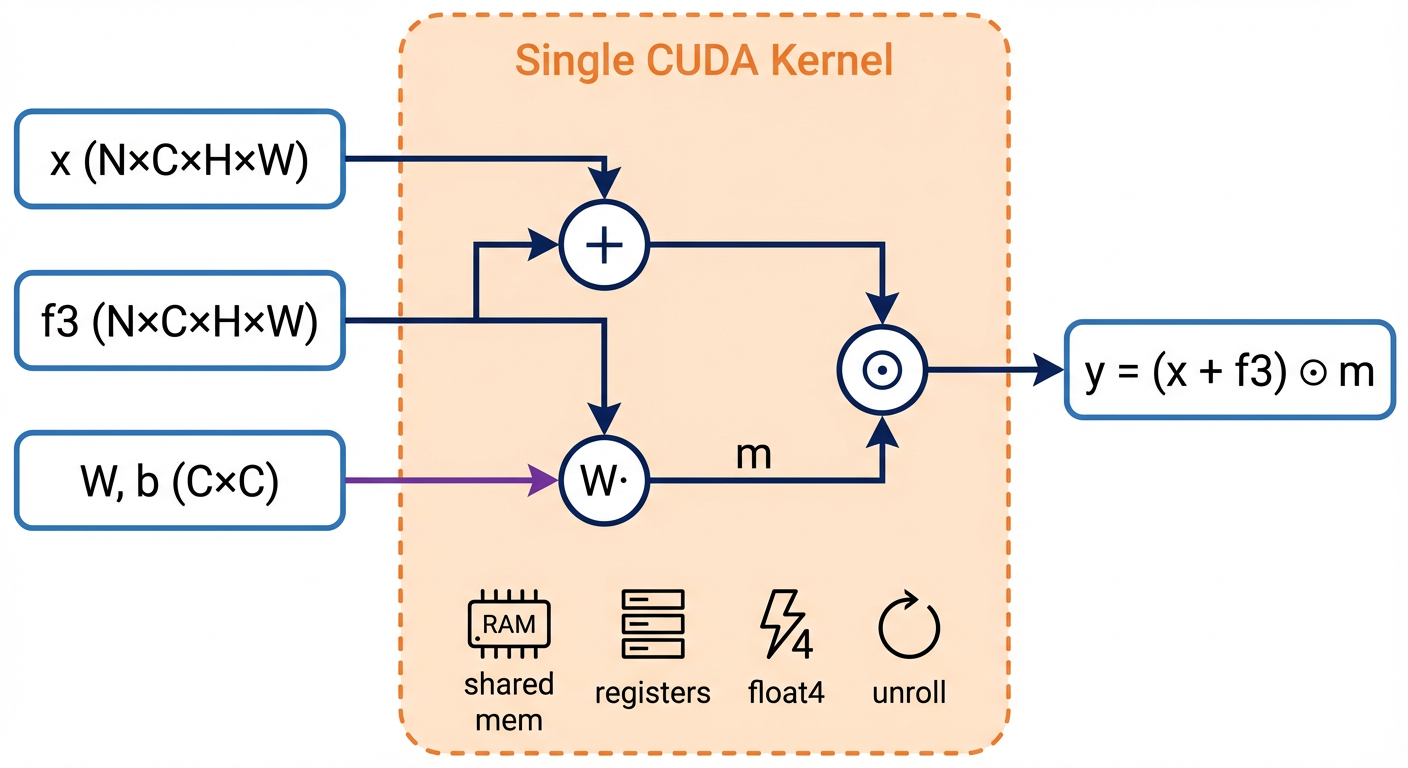}
  \caption{Team \textit{XiaomiMM}: \texttt{span\_attn\_op} fuses the $1\!\times\!1$ attention convolution,
    element-wise addition, and element-wise multiplication into a single
    CUDA kernel, eliminating $3\!\times$ redundant DRAM round-trips.}
  \label{fig:spanv2_op}
\end{figure}

\subsubsection{Model Architecture}

SPANV2 is a pure-CNN model (0.139\,M parameters, 32 feature channels)
comprising five sequential SPABV2 blocks, a near-pixel upsampling branch,
and a depthwise-separable feature fusion head.
The overall pipeline is shown in \cref{fig:spanv2_arch} and proceeds as
follows.

\begin{enumerate}[leftmargin=*, label=\textbf{\arabic*.}]
  \item \textbf{Near-pixel branch.}
    A depthwise convolution \texttt{conv\_near} maps the input
    $\mathbf{x}\!\in\!\mathbb{R}^{3\times H\times W}$ to
    $\mathbf{F}_\text{near}\!\in\!\mathbb{R}^{48\times H\times W}$
    ($3\times s^2$ channels, $s=4$). Its weights are initialized as a
    pixel-repeat operator so that each sub-channel initially copies the
    center-pixel value of the corresponding input channel, thereby providing
    a strong low-frequency prior at no additional inference cost.

  \item \textbf{Deep feature extraction.}
    Five SPABV2 blocks are applied sequentially.
    Block~1 projects the 3-channel input to 32 feature channels, while
    blocks 2--5 maintain 32 channels throughout.
    The output of block~5 is $\mathbf{F}_5\!\in\!\mathbb{R}^{32\times H\times W}$.

  \item \textbf{Feature fusion.}
    $\mathbf{F}_\text{near}$ (48\,ch) and $\mathbf{F}_5$ (32\,ch) are
    concatenated along the channel axis, yielding an
    $80\!\times\!H\!\times\!W$ feature tensor.
    A depthwise convolution (80\,ch$\to$80\,ch) followed by a pointwise
    convolution (80\,ch$\to$48\,ch) refines this joint representation.

  \item \textbf{Reconstruction.}
    PixelShuffle($\times4$) rearranges the 48-channel feature tensor into the
    final $3\!\times\!4H\!\times\!4W$ output.
\end{enumerate}

\subsubsection{SPABV2 Block}
\label{sec:spab}

The SPABV2 block extends the SPAB block of SPAN~\cite{span}.
Given input $\mathbf{x}\!\in\!\mathbb{R}^{C\times H\times W}$:
\begin{align}
  \mathbf{f}_1 &= \text{ReLU}(\text{Conv}_{3\times3}(\mathbf{x})),
  \label{eq:f1}\\
  \mathbf{f}_2 &= \text{ReLU}(\text{Conv}_{3\times3}(\mathbf{f}_1)),
  \label{eq:f2}\\
  \mathbf{f}_3 &= \text{Conv}_{3\times3}(\mathbf{f}_2),
  \label{eq:f3}\\
  \mathbf{m}   &= \text{Conv}_{1\times1}(\mathbf{f}_3),
  \label{eq:attnmap}\\
  \mathbf{y}   &= (\mathbf{x} + \mathbf{f}_3)\odot\mathbf{m}.
  \label{eq:attn}
\end{align}
The only structural change from SPAN's SPAB lies in \cref{eq:attnmap,eq:attn}:
a dedicated $1\!\times\!1$ convolution produces $\mathbf{m}$ from $\mathbf{f}_3$,
which then gates the residual $(\mathbf{x}+\mathbf{f}_3)$.
This design removes the restrictions of the original parameter-free attention:
$\mathbf{m}$ is no longer limited to non-negative element-wise interactions
between fixed intermediate features, and can instead take negative values and
perform cross-channel mixing, while adding only $C^2$ parameters per block and
requiring neither softmax nor normalization.
can now take negative values and perform cross-channel mixing, while adding only
$C^2$ parameters per block and requiring neither softmax nor normalization.

\subsubsection{\texttt{span\_attn\_op}: Fused CUDA Operator}
\label{sec:op}

\texttt{span\_attn\_op} fuses the three operations in
\cref{eq:attnmap,eq:attn}---Conv$_{1\times1}$, element-wise addition, and
element-wise multiplication---into a \emph{single CUDA kernel}:
\begin{equation}
  y_{c,h,w} =
  \bigl(x_{c,h,w} + f_{3,c,h,w}\bigr)
  \cdot
  \Bigl(b_c + \textstyle\sum_{i}W_{c,i}\,f_{3,i,h,w}\Bigr),
  \label{eq:fused}
\end{equation}
where $W\!\in\!\mathbb{R}^{C\times C}$ and $b\!\in\!\mathbb{R}^C$ are the
attention-convolution parameters.
Key optimization strategies include shared-memory weight caching,
\texttt{float4} vectorized loads, register-file feature caching, loop unrolling,
and channel-specialized kernels for $C\!\in\!\{16,28,32,48,52\}$.

\subsubsection{Near-Pixel Upsampling Branch}

\texttt{conv\_near} is a $3\!\times\!3$ depthwise convolution
($3$\,ch\,$\to$\,$48$\,ch, groups$\!=\!3$) initialized so that only the
center weight is non-zero:
\begin{equation}
  W_\text{near}[c\cdot s^2+k,\;0,\;1,\;1] = 1,\quad
  c\!=\!0,1,2;\;\;k\!=\!0,\dots,s^2\!-\!1,
\end{equation}
with all other weights set to zero ($s\!=\!4$).
At initialization, this degenerates into a \emph{$1\!\times\!1$
convolution}, so \texttt{conv\_near} followed by PixelShuffle($\times4$)
is exactly equivalent to \textbf{nearest-neighbor upsampling}---a principled
inductive bias, since low-frequency content (which NN reproduces exactly)
accounts for the majority of signal energy in natural images.
The weights remain trainable, allowing the branch to incorporate local context
beyond the center pixel during optimization.

%% =====================================================================
\subsubsection{Training Strategy}
%% =====================================================================

Training proceeds in two stages.

\textbf{Stage~1: Multi-scale pre-training.}
\begin{itemize}[leftmargin=*, topsep=2pt, itemsep=1pt]
  \item \textbf{Training data:} FD2K (Flickr2K + DIV2K); LR images
    generated by bicubic $\times4$ downsampling.
  \item \textbf{Patch sizes:} multi-shape HR crops sampled from eight
    resolutions ranging from $256\!\times\!256$ to $512\!\times\!512$
    (square, landscape, portrait), providing scale and aspect-ratio
    diversity.
  \item \textbf{Loss function:} $L_1$ pixel loss (weight $1.0$) +
    frequency-domain FFT loss (weight $0.05$).
  \item \textbf{Optimizer:} AdamW ($\beta_1=0.9$, $\beta_2=0.99$,
    weight decay $10^{-4}$); initial LR $10^{-3}$.
  \item \textbf{Scheduler:} cosine annealing over $10^6$ iterations,
    $\eta_\text{min}=10^{-6}$.
  \item \textbf{Batch size:} 8 per GPU.
  \item \textbf{Augmentation:} random horizontal flip and
    $90^\circ$ rotation.
  \item \textbf{EMA:} exponential moving average of weights
    ($\text{decay}=0.999$) throughout training.
\end{itemize}

\textbf{Stage~2: Fine-tuning.}
\begin{itemize}[leftmargin=*, topsep=2pt, itemsep=1pt]
  \item \textbf{Initialization:} weights from the Stage~1 checkpoint
    ($10^6$ iterations).
  \item \textbf{Patch size:} fixed $512\!\times\!512$ HR crops.
  \item \textbf{Loss function:} MSE loss (weight $5.0$) + gradient
    loss (weight $3.0$).
  \item \textbf{Optimizer:} AdamW ($\beta_1=0.9$, $\beta_2=0.99$,
    weight decay $10^{-4}$); initial LR $5\!\times\!10^{-4}$.
  \item \textbf{Scheduler:} cosine annealing with one warm restart
    at $6\!\times\!10^5$ iterations (restart weight $0.5$),
    for a total of $10^6$ iterations, $\eta_\text{min}=10^{-6}$.
  \item \textbf{Batch size:} 8 per GPU.
  \item \textbf{EMA:} same as in Stage~1 ($\text{decay}=0.999$).
\end{itemize}

%% file: teams/team01_BOE_AIoT/main.tex
\subsection{BOE\_AIoT}
Team \textit{BOE\_AIoT} select SPANF \cite{ren2025tenth} as the base model, as it represents the latest iteration of a model that has achieved top rankings for several consecutive years. Our goal is to obtain a faster and smaller model without significantly compromising image restoration performance. To this end, Team \textit{BOE\_AIoT} prune the final layer of the model, reducing the number of channels from 32 to 20, while leaving all other modules unchanged. To recover model performance, Team \textit{BOE\_AIoT} employ knowledge distillation for fine-tuning, enabling the image restoration capability to meet the required threshold for the competition. Team \textit{BOE\_AIoT} refer to this model as the Pruned and Distilled Super-resolution Model, abbreviated as PDS.

\noindent\textbf{Training Details}. After pruning the SPANF, Team \textit{BOE\_AIoT} train it on the DIV2K and LSDIR datasets, cropping the patch size to 256. The random rotation and flip are configured for data augmentation. The Adam optimizer and the L1 loss function are adopted to optimize the models, and the mini-batch size is set to 16.

%% file: teams/team16_PKDSR/main.tex
\subsection{PKDSR}
The solution is mainly inspired by SPAN \cite{wan2024swift}, and SPANF \cite{ren2025tenth}, which achieved competitive performance in the past NTIRE Efficient Super-Resolution Challenge.
The goal of this work is to further reduce the computational cost of the model, including FLOPs, parameters, and inference time, while maintaining the reconstruction quality in terms of PSNR and SSIM as much as possible.

To achieve this goal, we prune the tail feature channels of SPANF, reducing the number of channels from 32 to 24. 
Instead of pruning the channels directly, we perform the pruning in two stages (32→28→24), which makes the training process more stable and reduces the performance degradation after pruning.

Specifically, we first remove four channels from the pretrained model and then retrain the pruned model using knowledge distillation.
The unpruned model serves as the teacher model, while the pruned model acts as the student model.
Two types of losses are used during training: a knowledge distillation loss (KDLoss) and a reconstruction loss (RecLoss).
The reconstruction loss measures the difference between the output of the student model and the ground-truth high-resolution image, while the knowledge distillation loss enforces consistency between the outputs of the teacher and student models.
We empirically observe that the outputs of the teacher and pruned student models remain structurally similar after pruning; therefore, the distillation loss helps the student model relearn the features that are lost during the pruning process.

The overall training objective is defined as

\[
L = \alpha_1 L_{KD} + \alpha_2 L_{Rec},
\]
where $\alpha_1$ and $\alpha_2$ are balancing coefficients.

\noindent\textbf{Training Details.}
The model is trained on DIV2K \cite{agustsson2017ntire} and LSDIR \cite{lilsdir}.
We follow the training pipeline described in SPANF \cite{ren2025tenth} as the starting point of this method. 
Then, we prune the tail part of the model pipeline by gradually reducing the number of tail feature channels from 32 to 28 and finally to 24 through the following two steps:

\begin{enumerate}
    \item First, we prune the pretrained model by removing 4 channels from the tail features.
    The original unpruned model is used as the teacher model to guide the pruned student model to recover its performance.
    The learning rate is set to $1\times10^{-4}$, which is $\dfrac{1}{5}$ of the pretraining learning rate.
    Adam is used as the optimizer with $\beta_1=0.9$ and $\beta_2=0.99$.
    We adopt CosineAnnealingRestartLR as the learning rate scheduler, which gradually decreases the learning rate to $1\times10^{-7}$.
    The training process runs for a total of 50k iterations.
    The loss functions, KDLoss and RecLoss, are both implemented using L1 loss. And $\alpha_1$ and $\alpha_2$ are both set to be 1.

    \item Second, we further prune the model from 28 channels to 24 channels using the same settings.
    The only difference is that the number of iterations is increased to 100k to ensure sufficient performance recovery.
\end{enumerate}

%% file: teams/team04_ZenoSR/main.tex
\subsection{ZenoSR}
\begin{figure*}[t]
    \centering
    \includegraphics[width=0.9\linewidth]{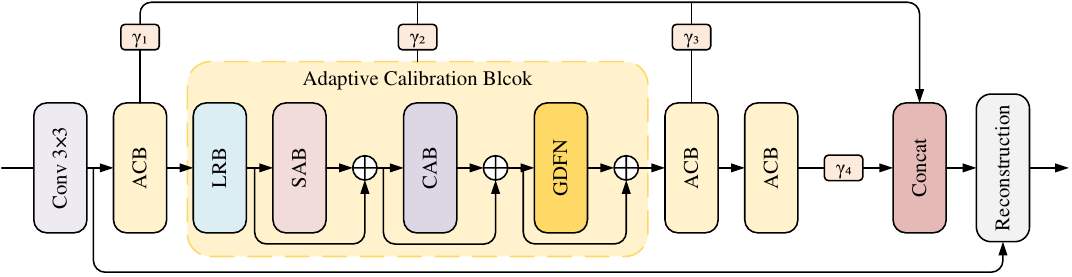}
    \caption{Team \textit{ZenoSR}: The overall architecture of Adaptive Calibration Network(ACN)}
    \label{fig_zenosr}
\end{figure*}
\noindent \textbf{Method Description.}
The architecture of their network is shown in Figure \ref{fig_zenosr}, which is inspired by previous studies such as DAN \cite{ren2025tenth} and SMFANet \cite{smfanet}. They propose an Adaptive Calibration Network (ACN). More specifically, they build upon the DAN framework by reducing the backbone depth and introducing a lightweight adaptive fusion strategy for efficient reconstruction. They preserve the hybrid dual-attention design of DAN, where each block combines local handling, spatial attention, and channel attention for complementary feature extraction, while reducing the number of stacked blocks from six to four to better satisfy the strict parameter and FLOPs budgets. To compensate for the reduced depth, they further propose a learnable stage-wise calibration scheme in the reconstruction pathway. Instead of directly concatenating intermediate features from different stages, they first reweight each stage output with an independent channel-wise learnable factor and then aggregate them for feature fusion. In this way, the network can automatically adjust the contribution of shallow and deep features according to their reconstruction importance, leading to more effective multi-level feature utilization with negligible overhead.

\noindent \textbf{Training strategy.}
The proposed ACN consists of 4 ACBs with 18 feature channels. They adopt a five-stage training strategy.

\begin{itemize}
    \item[1.] In the first stage, the model is trained from scratch on the 800 training images of DIV2K \cite{DIV2K} and the first 10K images of LSDIR \cite{lilsdir}. The cropped HR image size is set to 288, and the batch size is set to 64. The model is trained by minimizing the combination of L1 loss and frequency loss with the Adam optimizer. The initial learning rate is set to 2e-3 and is decayed at [200k, 400k, 600k, 700k]. The total number of iterations is 800k.

    \item[2.] In the second stage, the training datasets remain unchanged. The cropped HR image size and batch size are set to 288 and 64, respectively. The model is optimized by minimizing the combination of L1 loss and frequency loss. The initial learning rate is set to 2e-3 and is decayed at [200k, 400k, 600k, 700k]. The total number of iterations is 800k.

    \item[3.] In the third stage, the model is further trained on DIV2K and the entire LSDIR, while the cropped HR image size and batch size are adjusted to 384 and 64, respectively. The model is optimized by minimizing the combination of L1 loss and frequency loss. The initial learning rate is set to 1e-3 and is decayed at [200k, 400k, 600k, 700k]. The total number of iterations is 800k.

    \item[4.] In the fourth stage, the model is fine-tuned on the same datasets as the third stage using only the frequency loss. The cropped HR image size and batch size are set to 384 and 64, respectively. The initial learning rate is set to 5e-4 and is decayed at [200k, 400k, 600k, 700k]. The total number of iterations is 800k.

    \item[5.] In the final stage, the model is fine-tuned on DIV2K and the entire LSDIR with the L2 loss. The cropped HR image size and batch size are set to 512 and 64, respectively. The initial learning rate is set to 2e-4 and is decayed at [200k, 400k, 600k, 700k]. The total number of iterations is 800k.
\end{itemize}

%% file: teams/team05_CUIT_HTT/main.tex
\subsection{CUIT\_HTT}
\begin{figure*}[!t]
  \centering
  \includegraphics[width=\textwidth]{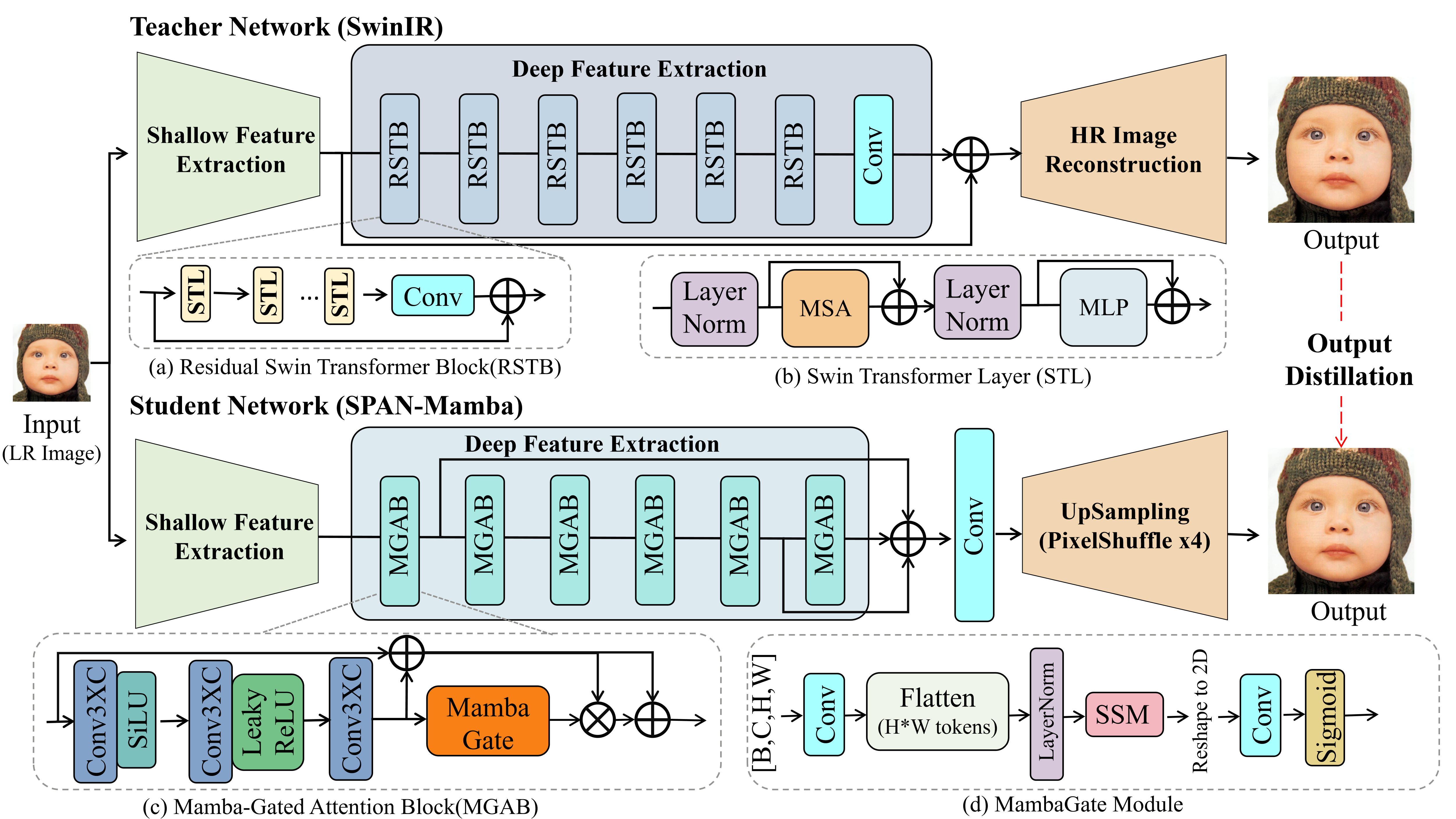}
  \caption{\textit{Team CUIT\_HTT}: Overall architecture of the proposed MambaGate-SR.}
  \label{fig:figureteam05}
\end{figure*}
\noindent\textbf{Method Description.} They propose \textbf{MambaGate-SR}, an efficient super-resolution method that enhances a lightweight SPAN-style backbone \citep{wan2024swift} with a Mamba-based residual gating mechanism. As shown in Fig.~\ref{fig:figureteam05}, the overall framework follows a teacher-student paradigm, where a SwinIR model \citep{SwinIR} is adopted as the teacher and a lightweight CNN-based model is used as the student. Different from recent Mamba-based restoration methods that directly replace the backbone with stacked state space blocks, their method preserves the efficient convolutional structure of SPAN and introduces Mamba only into a lightweight gating branch. In this way, MambaGate-SR enables global dependency modeling with limited computational overhead and is well-suited for efficient SR. The student network consists of shallow feature extraction, deep feature extraction, and HR reconstruction with a PixelShuffle ($\times4$) upsampling module. The deep feature extraction stage is built with stacked \textbf{Mamba-Gated Attention Blocks (MGABs)}. Each MGAB contains several re-parameterizable Conv3XC layers for efficient local feature extraction, followed by a lightweight \textbf{Mamba Gate} for global context modulation. Specifically, the intermediate feature map is first projected by a $1\times1$ convolution and reshaped into a 1D token sequence. The sequence is then normalized and processed by a Mamba-based State Space Model (SSM) \citep{mamba} to capture long-range spatial dependencies. The output tokens are reshaped back to the spatial domain and transformed into a residual gate, which is used to adaptively recalibrate the convolutional features. Therefore, the proposed design effectively combines efficient local CNN modeling and lightweight global Mamba modeling within a unified SR framework.\par
\noindent\textbf{Training Details.}
They use the SwinIR teacher to supervise the student with a hybrid distillation strategy. The total loss is defined as
\begin{equation}
    \mathcal{L}_{total} = \mathcal{L}_{pixel} + \lambda_1 \mathcal{L}_{kd\_out} + \lambda_2 \mathcal{L}_{kd\_edge},
\end{equation}
where $L_{\text{pixel}}$ is the L1 reconstruction loss, $L_{\text{kd\_out}}$ is the L1 output distillation loss between the student and teacher predictions, and $L_{\text{kd\_edge}}$ is a Laplacian-based edge distillation loss for enhancing high-frequency details. They set $\lambda_1=0.2$ and $\lambda_2=0.1$. The model is trained on a mixed DIV2K+LSDIR dataset  \citep{DIV2K,lilsdir} with random horizontal flips and rotations. The HR patch size is $256\times256$, the batch size is 16, and Adam is adopted with $\beta_1=0.9$ and $\beta_2=0.99$. The initial learning rate is $2\times10^{-4}$ and is decayed by a factor of 0.5 at 200k iterations using MultiStepLR. The total number of training iterations is 300k. In addition, an EMA strategy with decay 0.999 is employed to improve training stability.

%% file: teams/team06_HAESR/main.tex
\subsection{HAESR}

\begin{figure} 
    \centering 
    \includegraphics[width=1\linewidth]{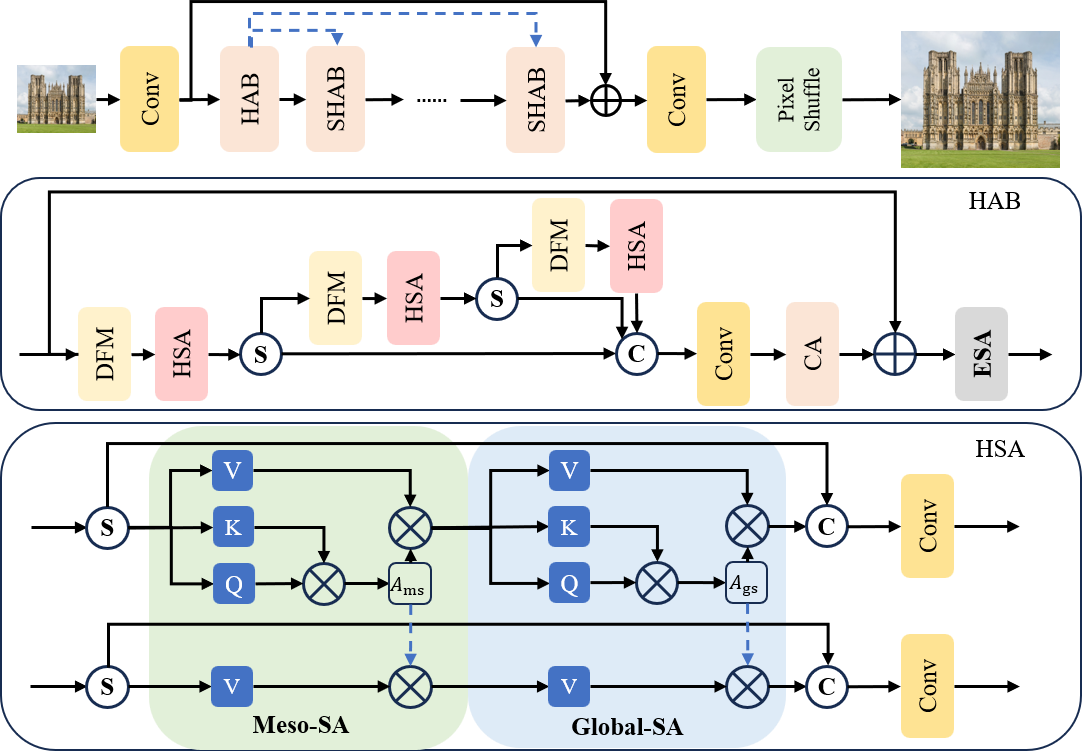} 
    \caption{\textit{Team HAESR}: The architecture of HAESR} \label{fig:network} 
\end{figure}

\textbf{Method Description.} They proposed HAESR, as shown in Fig.\ref{fig:network}. The work is inspired by ASID \cite{park2025efficient}, LKMN \cite{hu2025large}, OmniSR \cite{wang2023omni}, and HAT \cite{chen2023activating}. HAESR consists of three parts, shallow feature extraction, deep feature extraction and reconstruction module. Specifically, the deep feature extraction part is composed of one Hybrid Attention Block (HAB) and multiple Shared Hybrid Attention Blocks (SHABs). The HAB is used to extract hybrid-attention features and generate shared attention maps, while the SHABs reuse the previously generated attention maps for efficient feature refinement.

The HAB consists of a Dual-branch Feature Mixer (DFM) \cite{hu2025large} and a Hybrid Spatial Attention (HSA) module. The DFM is used to aggregate complementary local and contextual information, while the HSA further enhances the features through Meso-SA and Global-SA. Inspired by OmniSR~\cite{wang2023omni} and HAT~\cite{chen2023activating}, the proposed spatial attention HSA adopts extended local sampling and structured global sampling to achieve efficient local-global feature modeling. Unlike HAB, SHAB does not generate new attention maps. Instead, it directly reuses the attention maps produced by the preceding HAB and only performs lightweight feature transformation and aggregation. In this way, the first block establishes the attention relationships, while the following blocks refine features based on the shared attention priors, which effectively reduces redundant computation and improves network efficiency.

\noindent\textbf{Training Details.} The model is trained on DIV2K \cite{agustsson2017ntire} and LSDIR \cite{lilsdir}. First, the model is trained for a total of 500k iterations by minimizing L1 loss with AdamW optimizer ($\beta_1= 0.9$, $\beta_2=0.99$) . The training HR patch size is set to 256×256 with data augmentation and the batch size is set to 32. The initial learning rate is set to 5e-4 and halved at 250k, 400k, 450k, 475k iterations. Then, the training HR patch size is set to 384×384 and the batch size is set to 32. The model is further trained for 300k iterations. The initial learning rate is set to 5e-5 and halved every 100k iterations. The proposed  method is implemented using the PyTorch framework on a single NVIDIA RTX 4090 GPU.

%% file: teams/team09_IN2GM/main.tex
\subsection{IN2GM}

\begin{figure}[!t]
    \centering
    \includegraphics[width=1\linewidth]{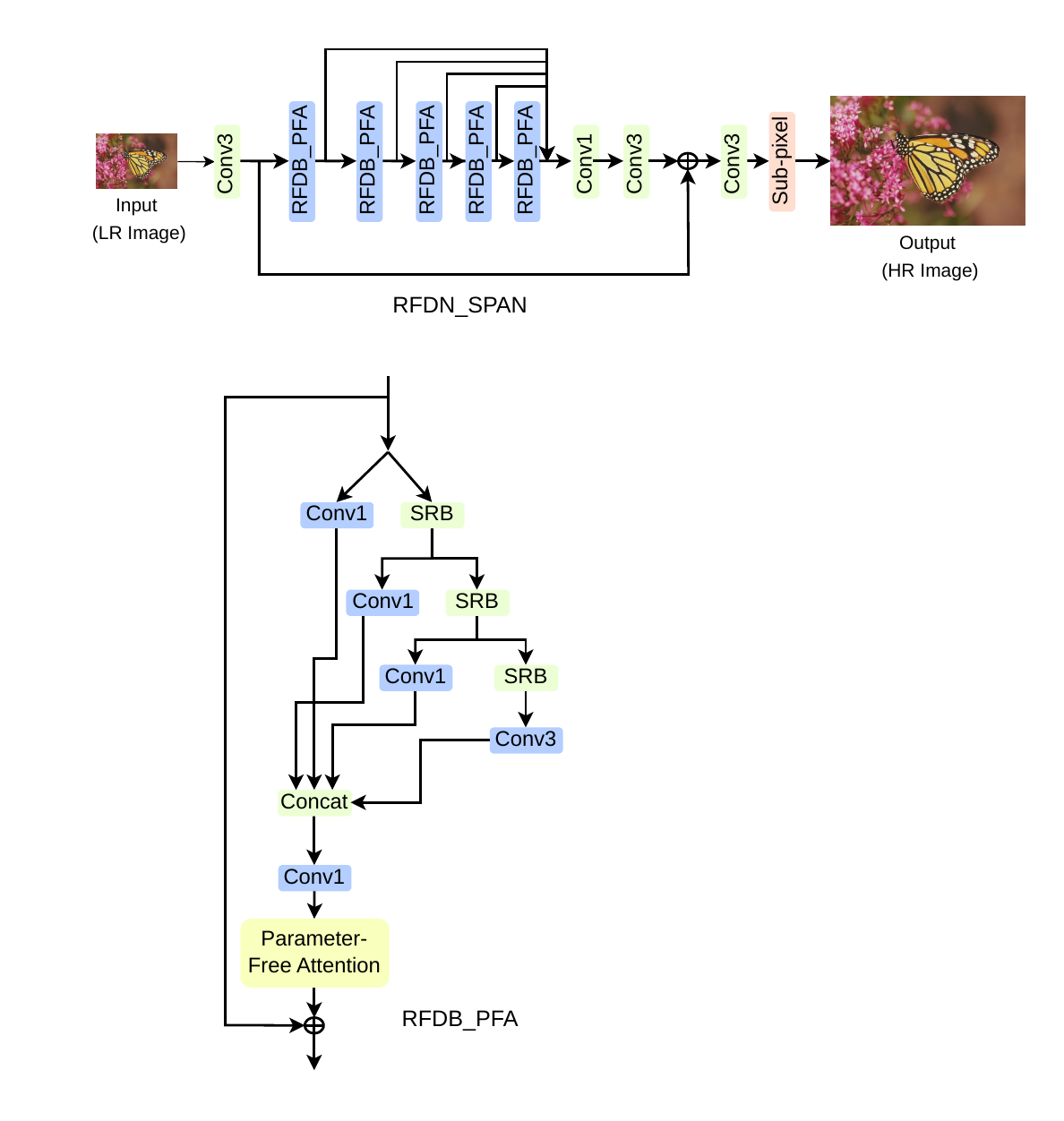}
    \caption{Team \textit{IN2GM}: Overview of the proposed RFDN\_SPAN architecture.}
    \label{fig:RFDN_SPAN}
\end{figure}

\paragraph{Method.}
They propose \textbf{RFDN\_SPAN}, an efficient model built upon RFDN~\cite{liu2020residualfeaturedistillationnetwork}. The overall architecture is illustrated in Fig.~\ref{fig:RFDN_SPAN}. The core modification is a redesign of the residual feature distillation block (RFDB), where the enhanced spatial attention (ESA) module is replaced with a parameter-free attention mechanism inspired by SPAN~\cite{wan2024swift}. This attention computes spatial importance directly from intermediate features without introducing additional parameters, enabling effective feature modulation while reducing computational overhead. As a result, the proposed model achieves lower parameter count and FLOPs while maintaining competitive reconstruction performance.

\paragraph{Training details.}
The model uses 46 feature channels and is trained on the DIV2K and LSDIR datasets with random horizontal flips and $90^\circ$ rotations. They adopt the Adam optimizer with $\beta_1=0.9$ and $\beta_2=0.999$. Training is performed in three progressive stages with increasing patch sizes (256, 384, 512) and a mini-batch size of 64. The initial learning rate is set to $5\times10^{-4}$ for the first two stages and $2\times10^{-4}$ for the final stage, with a decay factor of 0.5 applied every $2\times10^5$ iterations.

%% file: teams/team10_Sunflower/main.tex
\subsection{Sunflower}
\begin{figure*}[!t]  
    \centering
    \includegraphics[width=\textwidth]{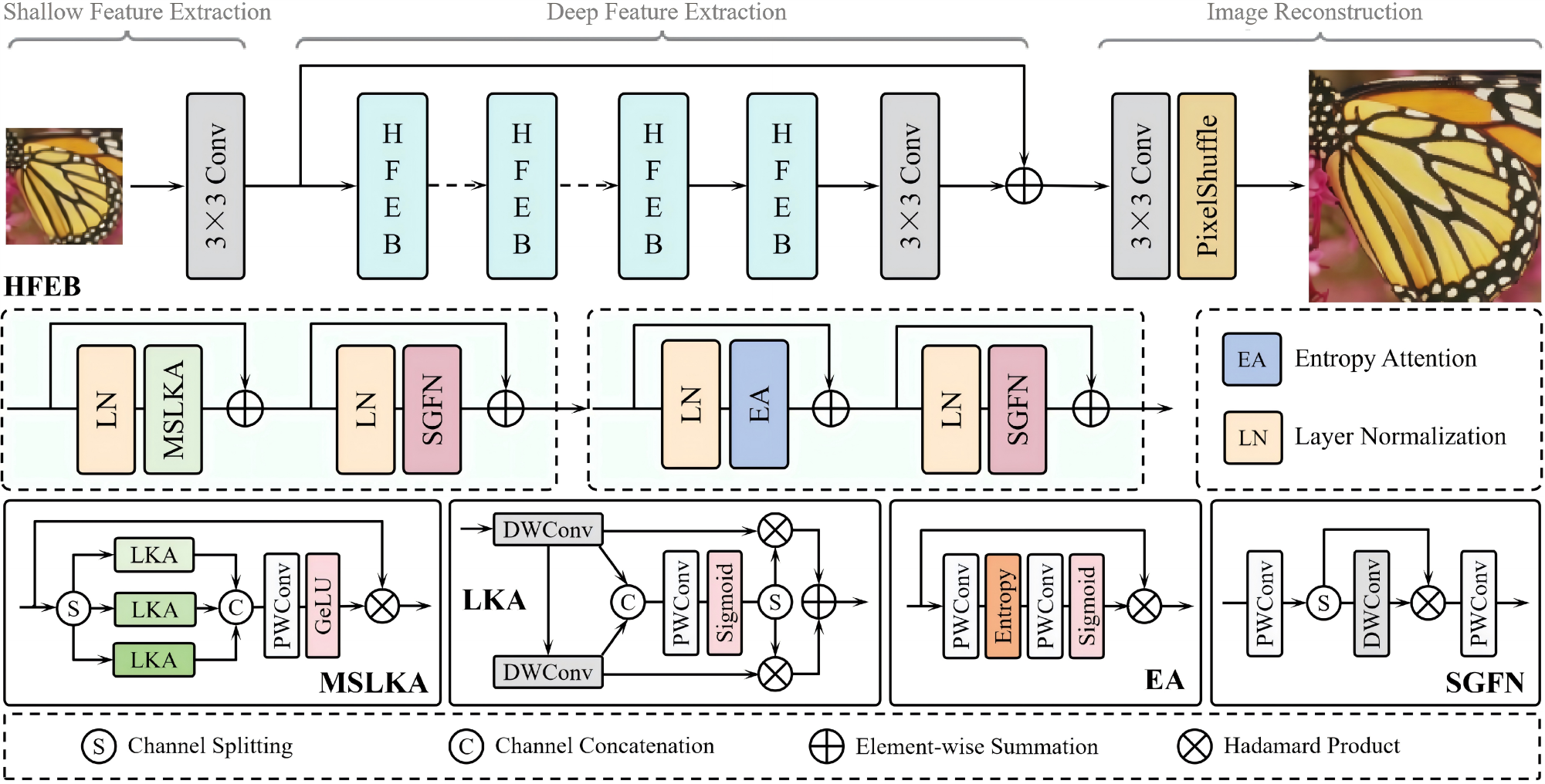}
    \caption{Team \textit{Sunflower}: The network architecture of HFENet}
    \label{fig:example}
\end{figure*}

\textbf{Method Description.} 
The architecture of the proposed network is depicted in Fig.\ref{fig:example}, which is inspired by previous studies such as LSKNet \cite{li2023lsknet} and EARFA \cite{zhao2024efficient}. They propose a Hierarchical Feature Enhancement Network (HFENet) for efficient single-image super-resolution. More specifically, they build upon the efficient feature extraction framework by constructing a cascaded processing branch consisting of a Multi-Scale Large Kernel Attention (MSLKA) module and an Entropy Attention (EA) module to enhance the overall performance in a hierarchical way. Meanwhile, they replace the standard large kernel convolution with parallel multi-scale depth-wise dilated convolutions, and utilize the differential entropy conditioned on a Gaussian distribution instead of conventional channel pooling, for the sake of harvesting significant savings of the parameters while capturing highly informative representations.

Following the overall pipeline, the core of their HFENet lies in the Hierarchical Feature Enhancement Block (HFEB), which sequentially integrates spatial and channel enhancements. Inspired by LSKNet \cite{li2023lsknet}, the Multi-Scale Large Kernel Attention (MSLKA) module is designed to capture multi-scale spatial contexts. Given an input feature map, MSLKA evenly splits it into three groups along the channel dimension. To simulate a larger effective receptive field, each branch consists of two cascaded depth-wise convolutional (DWConv) layers. Let $(k_1, d_1)$ and $(k_2, d_2)$ represent the kernel size and dilation rate of the first and second DWConv layers, respectively. Their combinations can induce varying receptive fields to capture diverse spatial patterns. In their implementation, the first branch keeps $(k_1, d_1) = (3, 3)$ and $(k_2, d_2) = (5, 3)$. For the second one, $(k_1, d_1) = (3, 1)$ and $(k_2, d_2) = (7, 3)$. They set $(k_1, d_1) = (5, 1)$ and $(k_2, d_2) = (5, 5)$ for the last branch. The multi-scale features from these three branches are then concatenated and adaptively aggregated through point-wise convolutions and a Sigmoid activation, providing a comprehensive spatial attention map that recalibrates the original input.

After the multi-scale spatial features are enhanced by the MSLKA module, they are fed into the Entropy Attention (EA) module to further recalibrate the channel-wise information. To efficiently measure the information content of each channel, the EA module employs the differential entropy conditioned on a Gaussian distribution, following the design in EARFA \cite{zhao2024efficient}. Specifically, the input features first pass through a $1\times1$ convolutional layer to compress the channel dimension, reducing the overall computational overhead. Then, the spatial variance of each compressed channel is calculated to estimate its entropy value, which serves as a robust indicator of the information richness. The entropy calculation can be formulated as:

\begin{equation}
\mathrm{H}(\mathrm{z}) = \frac{1}{2}\ln(2\pi\sigma^2(\mathrm{z})).
\end{equation}

where $\sigma(\cdot)$ is the standard deviation, and $\ln(\cdot)$ signifies the napierian logarithm, and $\mathrm{H}(\mathrm{z})$ represents the differential entropy of the feature map conditioned on the Gaussian distribution. Subsequently, another $1\times1$ convolutional layer is utilized to restore the channel dimension, followed by a Sigmoid activation function to generate the normalized channel attention weights. Finally, these weights are applied to the original input features via element-wise multiplication, selectively emphasizing the channels with richer representations.

\textbf{Training Details.} 
They trained the proposed HFENet model on a mixed dataset composed of DIV2K \cite{agustsson2017ntire} and LSDIR \cite{lilsdir}. They set the number of feature channels to 27 and the number of HFEBs to 7. During the training process, the crop size of input image patches was set to $128\times128$ with a batch size of 16. They adopted the Adam optimizer with $\beta_1 = 0.9$ and $\beta_2 = 0.999$ to minimize the $L_1$ loss. The total training procedure underwent 500k iterations. The initial learning rate was set to $5\times10^{-4}$ and was halved at the 250k, 400k, 450k, and 475k iterations to ensure stable convergence. Training was completed using a single NVIDIA GeForce RTX 4090 GPU.

%% file: teams/team11_XuptSR/main.tex
\subsection{XuptSR}

\begin{figure*}[!t]
    \centering
    \includegraphics[width=1.0\textwidth]{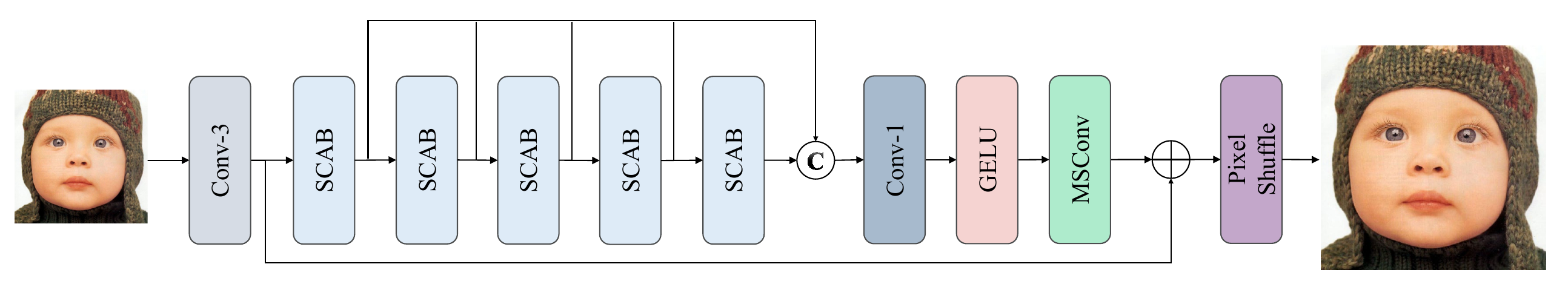}
    \caption{Team \textit{XuptSR}: The whole framework of Variance-Guided Spatial-Channel Context Interaction Network (VSCINet)}
    \label{VSCINet}
\end{figure*}

\begin{figure*}[!t]
    \centering
    \includegraphics[width=1.0\textwidth]{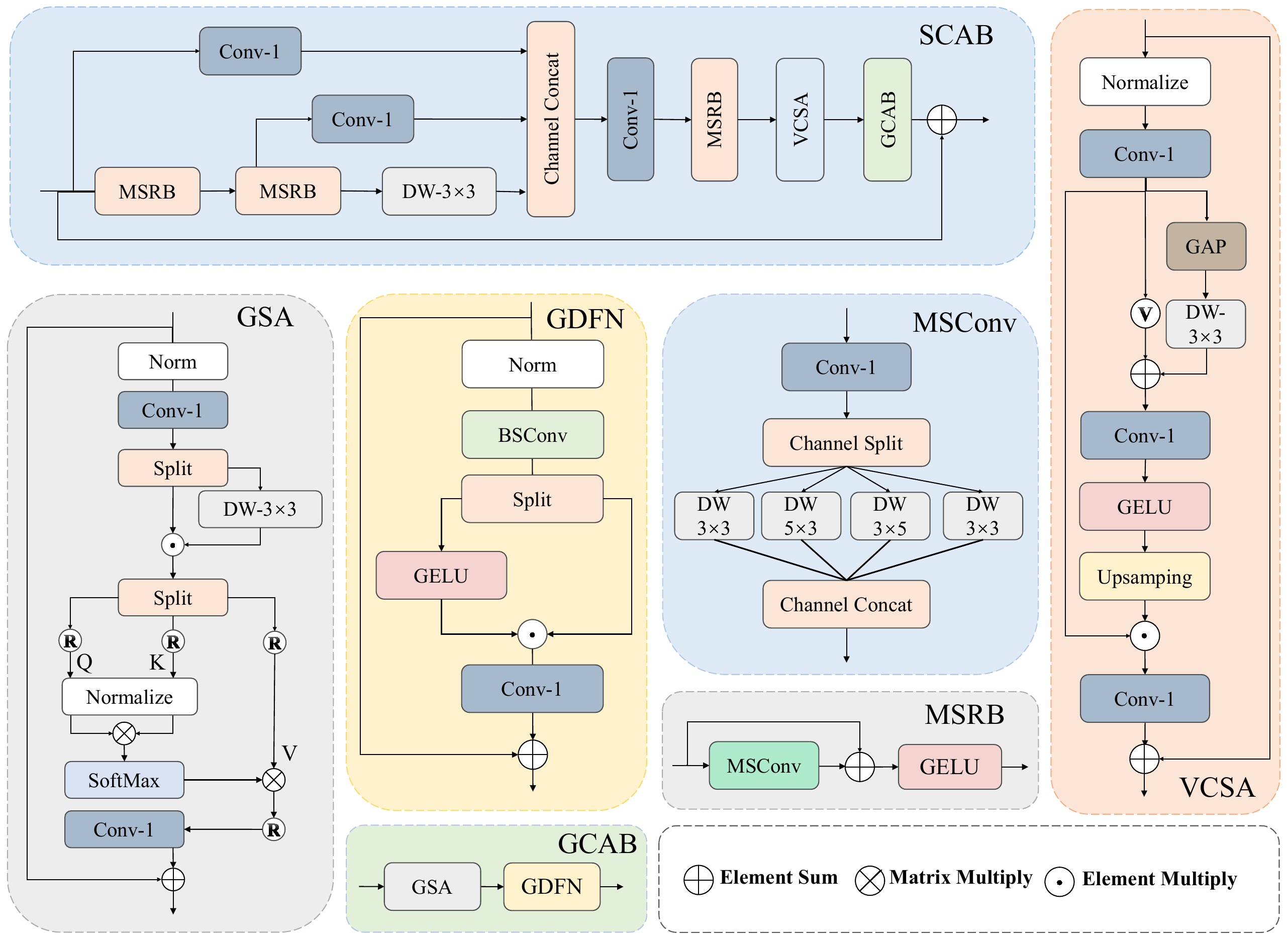}
    \caption{Team \textit{XuptSR}: The details of each component. SCAB: Spatial Channel Attention Block; GSA: Gated Self-Attention; GDFN: Gated Depthwise Feed-Forward Network; GCAB:Global Channel Attention Block; VCSA: Variance-guided Contextual Spatial Attention; MSConv: Multi-scale Convolution; MSRB: Multi-scale Residual Block}
    \label{details}
\end{figure*}
\noindent \textbf{Method Description.}
The XuptSR team proposed the Variance-Guided Spatial-Channel Context Interaction Network (VSCINet), as illustrated in Fig. \ref{VSCINet}. VSCINet follows a lightweight super-resolution framework similar to efficient reconstruction networks such as \cite{IMDN}. It consists of multiple attention-enhanced feature extraction blocks, a feature aggregation module, and an efficient reconstruction layer. By integrating spatial attention, channel attention, and multi-scale convolution, the network captures rich contextual information with low computational cost.

To improve feature representation, VSCINet introduces three key modules: Variance-guided Contextual Spatial Attention (VCSA), Global Channel Attention Block (GCAB), and Multi-Scale Convolution (MSConv), as shown in Fig. \ref{details}. VCSA combines contextual features and variance information to generate spatial attention maps, helping the network focus on informative regions. GCAB models channel dependencies through a gated self-attention mechanism and adopts a local chunking strategy at test time to reduce memory usage. MSConv employs parallel depthwise convolutions with different receptive fields to extract multi-scale spatial features and enhance feature diversity.

By stacking these attention-enhanced blocks, VSCINet progressively refines feature representations. The intermediate features are then concatenated and fused through a lightweight convolutional module, followed by a convolution layer and pixel shuffle upsampling for efficient high-resolution image reconstruction.

\begin{figure*}[!t]
    \centering
    \includegraphics[width=\linewidth]{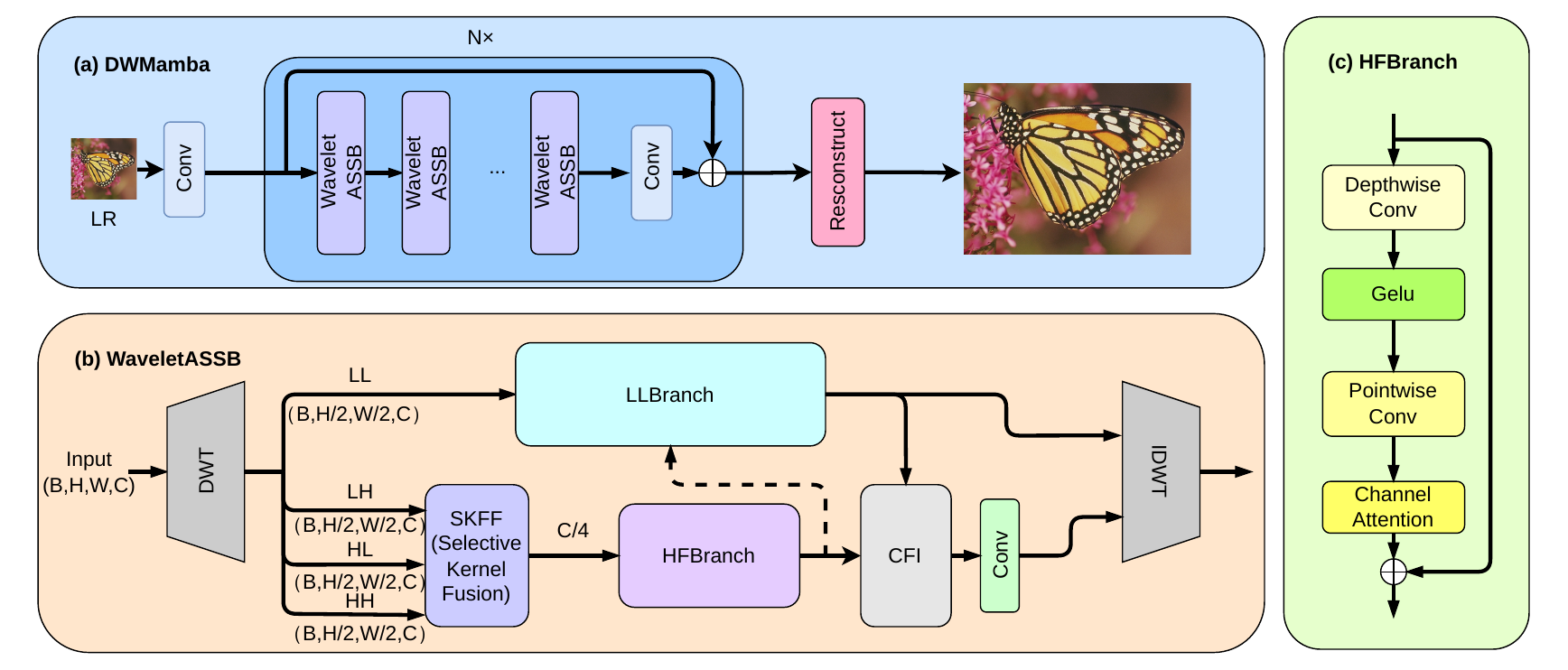}
    \caption{Team \textit{WMESR}: Overview of DWMamba.}
    \label{fig:dwmamba}
\end{figure*}

\noindent \textbf{Training strategy.}
The proposed VSCINet contains five stacked Spatial-Channel Attention Blocks (SCABs), where the number of feature channels is set to 20. The details of training steps are as follows:
\begin{enumerate}
    \item Pretraining on the DIV2K \cite{DIV2K} and the first 10K images of the LSDIR \cite{lilsdir}. HR patches of size 256 × 256 are randomly cropped from HR images, and the mini-batch size is set to 64. The model is trained by minimizing the L1 loss function with the Adam optimizer. The initial learning rate is set to $3 \times {10^{ - 3}}$ and halved at $\left\{ {100k, 500k, 800k,900k,950k} \right\}$-iteration. The total number of iterations is 1000k.
    \item Finetuning on 800 images of DIV2K and the Flickr2K \cite{Flickr2K}. HR patch size and  mini-batch size are set to 384 × 384 and 64, respectively. The model is fine-tuned by minimizing L2 loss function. The initial learning rate is set to $5 \times {10^{ - 4}}$ and halved at $\left\{ {100k, 500k, 800k,900k,950k} \right\}$-iteration. The total number of iterations is 1000k.
\end{enumerate}

%% file: teams/team12_WMESR/main.tex
\subsection{WMESR}
\noindent\textbf{Method}
To overcome the computational bottleneck of state-space models in high-resolution image restoration, they propose DualWave-Mamba (DWMamba), a lightweight, frequency-decoupled framework. It first extracts shallow features via a $3\times3$ convolution, passes them through a deep backbone of stacked Wavelet Attentive State Space Blocks (Wavelet ASSB), and reconstructs the output using a PixelShuffle module(as shown in Fig.~\ref{fig:dwmamba}).

The core Wavelet ASSB (illustrated in Fig.~\ref{fig:dwmamba}) uses a 2D Haar Discrete Wavelet Transform (DWT) to decompose features into one low-frequency (LL) and three high-frequency (HL, LH, HH) subbands. This lossless compression halves spatial resolution and reduces sequence length to one-quarter, significantly lowering memory usage and latency before feeding the subbands into a parallel dual-branch architecture.

The low-frequency branch adopts the MambaIRv2 \cite{guo2025mambav2} architecture to process the LL subband. They introduce a cross-branch injection mechanism where high-frequency features guide spatial token routing and modulate the Convolutional Positional Encoding (CPE), ensuring the Mamba module retains local texture perception.

For the inherently sparse high-frequency information, they fuse the HL, LH, and HH subbands via a Selective Kernel Feature Fusion (SKFF) module \cite{zamir2020mirnet}. To prevent this fast-converging CNN branch from dominating early training gradients, they compress its channels to one-quarter of the base dimension ($C/4$) before refining it with depthwise convolutions and channel attention.

A Cross-Frequency Interaction (CFI) module uses the processed LL features to generate a structural mask, selectively suppressing high-frequency noise. Finally, an Inverse Discrete Wavelet Transform (IDWT) merges the branches back to the original spatial resolution.

\noindent\textbf{Training Details}
To maximize efficiency without sacrificing quality, their compact DWMamba (0.674M parameters, 19.34G FLOPs) employs a progressive 4-stage Knowledge Distillation (KD) strategy. This pipeline transfers robust representations from a high-capacity Teacher to their lightweight Student network:
\begin{itemize}
    \item \textbf{Stage 1: Pre-training.} The Teacher is trained from scratch on DIV2K \cite{DIV2K} and LSDIR \cite{lilsdir} (L1 loss, $256\times256$ patch size, batch size 16, learning rate $2\times10^{-4}$, 300K iterations).
    \item \textbf{Stage 2: Large-Patch Fine-tuning.} To capture broader spatial contexts, the Teacher is fine-tuned with $384\times384$ patches (learning rate $5\times10^{-5}$, 150K iterations).
    \item \textbf{Stage 3: PSNR-Oriented Fine-tuning.} The Teacher is fine-tuned with an MSE loss to maximize PSNR directly (learning rate $1\times10^{-5}$, 50K iterations).
    \item \textbf{Stage 4: Knowledge Distillation.} The frozen Teacher supervises the compact Student (dimensions: 132$\rightarrow$48, blocks: 24$\rightarrow$8) for 250K iterations. The dual-supervision loss combines a 0.5-weighted L1 pixel loss (against Ground Truth) and a 1.0-weighted L1 distillation loss (against Teacher).
\end{itemize}
This pipeline effectively offsets performance drops from their aggressive lightweight design, ensuring an optimal balance between inference speed and restoration accuracy.

%% file: teams/team15_VARH-AI/main.tex
\subsection{VARH-AI}
\label{ssec:team15}
\begin{figure*}[!t]
  \centering
  \includegraphics[width=0.7\textwidth]{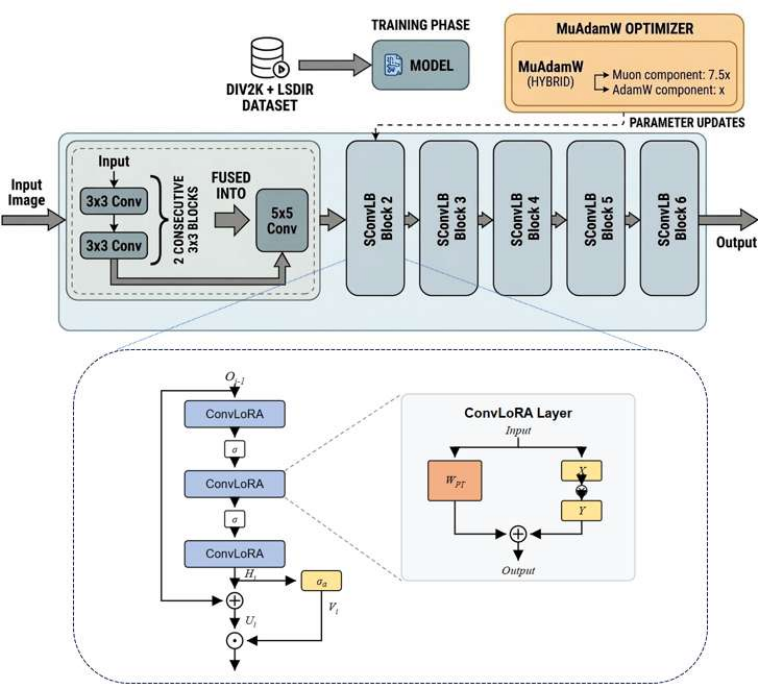}
  \caption{Team \textit{VARH-AI}: The proposed model architecture overview.}
  \label{fig:team15_arch}
\end{figure*}
\paragraph{Method Description}
The architecture proposed by VARH-AI is built upon the SPAN~\cite{wan2024swift} architecture from the 2024 challenge
winner and the DSCLoRA distillation framework from the 2025 winner
(Team EMSR)~\cite{ren2025tenth,chai2025dsclora}.
The solution focuses on three key contributions.
(1)~Exact Operator Fusion:
Sequential $3\!\times\!3$ convolutions within the baseline residual
connections are mathematically compacted into equivalent higher-order
kernels (e.g.\ $5\!\times\!5$), eliminating ${\sim}5$K parameters
without altering the output tensor distribution.
(2)~DSCLoRA Distillation:
Knowledge is transferred from a wider frozen teacher network to a
compressed student via Separable Convolutional LoRA Blocks (SConvLB)
and spatial affinity alignment.
(3)~MuAdamW Optimization:
A hybrid optimizer that routes rank-constrained convolutional parameters
through Newton-Schulz orthogonalization
(Muon)~\cite{ultralytics2026yolo26} while handling biases and
normalization scalars with standard AdamW, achieving faster and more
stable convergence on the heterogeneous DSCLoRA topology. The architecture follows the SPAN topology with 26 feature channels and
six feature extraction blocks.
The overall pipeline and SConvLB block detail are shown
in~\cref{fig:team15_arch}.

\begin{enumerate}[leftmargin=*, label=\textbf{\arabic*.}]
  \item \textbf{Frontend Operator Fusion.}
    They identify the linearly dependent cascade of the $3\!\times\!3$
    head convolution (\texttt{conv\_1}) into the first $3\!\times\!3$
    convolution of \texttt{block\_1} and mathematically compact
    them into a single equivalent $5\!\times\!5$ kernel.
    A similar partial contraction is applied at the backend between
    the $1\!\times\!1$ aggregation convolution (\texttt{conv\_cat}) and
    the $3\!\times\!3$ tail convolution (\texttt{conv\_2}).
    This zero-cost compression strips ${\sim}5$K parameters natively.

  \item \textbf{SConvLB Blocks.}
    Each of the six feature extraction blocks replaces the standard SPAB
    with a Separable Convolutional LoRA Block
    (SConvLB)~\cite{chai2025dsclora}.
    Within each block, three ConvLoRA layers inject low-rank pathways
    into the pre-trained depthwise convolutions.
    The LoRA decomposition factorizes the weight update as
    $\Delta W = BA$, where
    $A \!\in\! \mathbb{R}^{r k \times c_\text{in} k}$ and
    $B \!\in\! \mathbb{R}^{c_\text{out} k \times r k}$
    ($k$\,=\,kernel size), with scaling factor $\alpha / r$.
    At inference, these branches are re-parameterized and merged into
    the base weights, ensuring zero additional overhead.

  \item \textbf{Feature Aggregation.}
    Following SPAN, skip connections from the head, tail, the first
    block, and a late intermediate block are concatenated along the
    channel axis and refined by a $1\!\times\!1$ convolution before
    the upsampler.

  \item \textbf{Upsampling.}
    LoRA adaptation is extended beyond the main backbone to the pixel
    shuffle module and its preceding convolutional layer, ensuring
    low-rank updates permeate the entire upsampling pipeline.
\end{enumerate}

The final fused student model contains 125,992~parameters
and requires 8.22~GMACs at $\times4$ super-resolution.

\paragraph{Training Details} The architecture employs a student--teacher distillation pipeline with a
four-stage progressive curriculum.
Knowledge is transferred from a frozen 30-channel EMSR
teacher~\cite{ren2025tenth} to the 26-channel fused student via the
DSCLoRA pathways.
Spatial feature affinity matrices are aligned across multiple network
layers.
For a feature map $F_l \!\in\! \mathbb{R}^{B \times C \times H \times W}$
at the $l$-th layer, the affinity distillation objective is:
\begin{equation}
  \mathcal{L}_{\text{AD}} = \frac{1}{|\mathcal{A}|}
    \sum_{l=1}^{n} \bigl\| A_l^{S} - A_l^{T} \bigr\|_1,
  \label{eq:team15_affinity}
\end{equation}
where $A_l^{S}$ and $A_l^{T}$ denote the student and teacher spatial
affinity matrices computed after each SConvLB module. Training follows four progressive spatial patch resolutions, as
summarized in~\cref{tab:team15_curriculum}.
An auxiliary FFT loss is dynamically scheduled: its weight ramps
linearly from 0 to 0.5 during Stage~2, then decays back to 0 by the
end of Stage~3, providing transient frequency-domain supervision to
encourage high-frequency detail recovery before yielding to pure
pixel-accuracy optimization in the final stage.

\begin{table}[t]
  \centering
  \caption{DSCLoRA multi-stage curriculum schedule.}
  \label{tab:team15_curriculum}
  \small
  \begin{tabular}{ccccc}
    \toprule
    Stage & Patch Size & Batch & Iterations & FFT Weight \\
    \midrule
    1 & $128\!\times\!128$ & 64 & 200K & 0 \\
    2 & $256\!\times\!256$ & 64 & 200K & $0 \to 0.5$ \\
    3 & $384\!\times\!384$ & 64 & 200K & $0.5 \to 0$ \\
    4 & $512\!\times\!512$ & 64 & 200K & 0 \\
    \bottomrule
  \end{tabular}
\end{table}

\begin{itemize}[leftmargin=*, topsep=2pt, itemsep=1pt]
  \item \textbf{Training data:} Full DIV2K (800 images) + LSDIR
    (20\% ratio); LR images generated by bicubic $\times4$
    downsampling.
  \item \textbf{Loss function:} Charbonnier pixel loss (weight 1.0)
    + FFT frequency-domain loss (dynamically scheduled, peak weight
    0.5) + output distillation L1 loss (weight 0.1) + spatial affinity
    loss (weight 0.01).
  \item \textbf{Optimizer:} MuAdamW --- a hybrid optimizer inspired by
    MuSGD~\cite{ultralytics2026yolo26}.
    For rank-constrained convolutions ($\geq$\,2D weights), it applies
    a decoupled Newton-Schulz orthogonalization update (Muon) over 5
    iterative steps:
    \begin{align}
      A &= X_k X_k^\top, \label{eq:team15_ns1} \\
      X_{k+1} &= a\,X_k + (b\,A + c\,A^2)\,X_k, \label{eq:team15_ns2}
    \end{align}
    where $a\!=\!3.4445$, $b\!=\!{-4.7750}$, $c\!=\!2.0315$.
    Parameters are updated with scaled Nesterov momentum:
    \begin{equation}
      W_{t+1} = W_t\,(1 - \eta\lambda)
               - \eta_{\text{muon}}\,\mathrm{Ortho}(G_t).
      \label{eq:team15_update}
    \end{equation}
    Conversely, 1D parameters (biases, normalization scalars) are
    routed to a standard AdamW pipeline.
  \item \textbf{Muon learning rate} ($\eta_{\text{muon}}$):
    $1.5\!\times\!10^{-3}$ ($7.5\times$ the base AdamW rate of
    $2\!\times\!10^{-4}$).
  \item \textbf{AdamW betas} ($\beta_1, \beta_2$): $(0.9, 0.999)$.
  \item \textbf{Momentum} ($\mu$): $0.95$ with Nesterov enabled.
  \item \textbf{Weight decay} ($\lambda$): $1\!\times\!10^{-4}$.
  \item \textbf{Batch size:} 64 across all stages.
  \item \textbf{Total iterations:} 800K (200K per stage).
  \item \textbf{Augmentation:} Random horizontal/vertical flips and
    $90^\circ$ rotations (8-fold).
\end{itemize}

%% file: teams/team17_XSR/main.tex
\subsection{XSR}
\begin{figure*}[!t]
  \centering
 \includegraphics[width=1.0\linewidth]{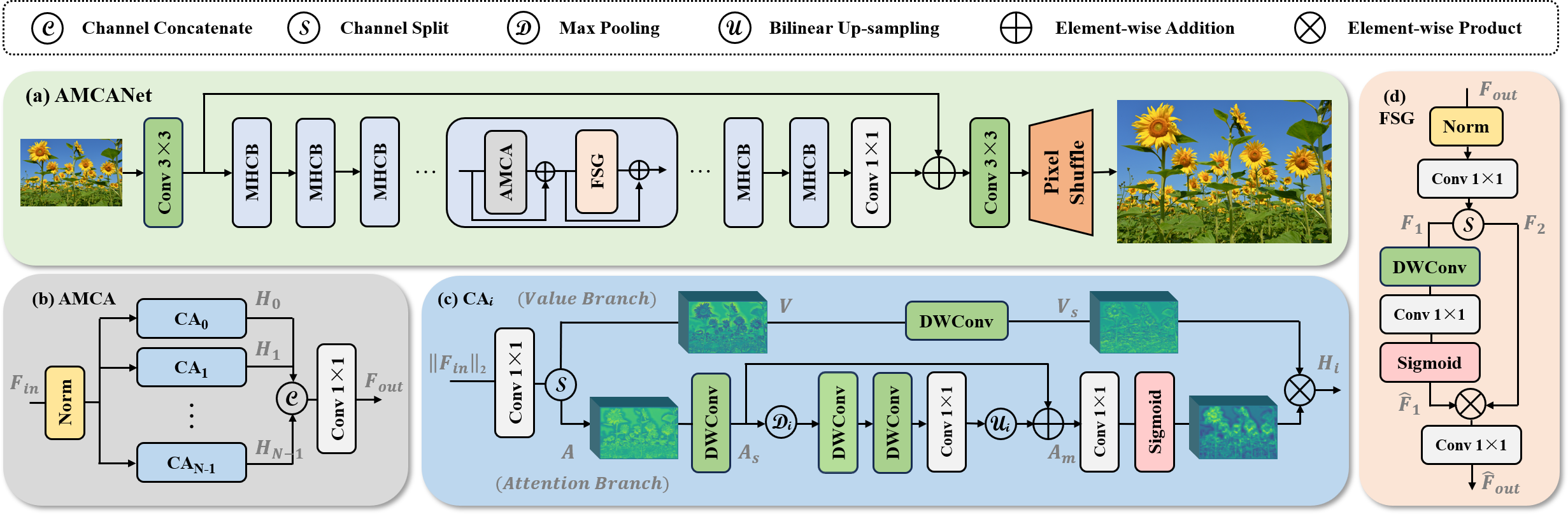}
  \caption{Team \textit{XSR}: Network architecture of the proposed AMCANet.}
  \label{fig:arch}
\end{figure*}

\textbf{Method.} They propose the Architecture-guided Multi-head Convolution Attention Network (AMCANet) for image super-resolution. The overall architecture is shown in Fig.~\ref{fig:arch}(a). The network first extracts shallow features via a $3\times3$ convolution, which are then processed by cascaded Multi-head Convolution Blocks (MHCBs). Each MHCB integrates an Architecture-guided Multi-head Convolution Attention (AMCA) module and a Feature Selective Gate (FSG) module to extract hierarchical features, formulated as:
\begin{equation}
    \begin{split}
        &F_{mid} = \text{AMCA}(F_{in})+F_{in},\\
        &F_{out} = \text{FSG}(F_{mid})+F_{mid}.
    \end{split}
\end{equation}
To efficiently model both local details and non-local contexts, the AMCA module employs multiple parallel Convolution Attention (CA) heads with different downsampling rates. For the $i$-th head, the normalized input feature is split into an attention branch $A$ and a value branch $V$. Their spatial features $A_s, V_s$ are extracted via depthwise convolutions. 
\begin{equation}
    \mathcal{F} = \text{Conv}_{1\times 1}(\text{DWConv}_{3\times 3}(\text{DWConv}_{3\times 3}(\cdot))).
    \label{eq:F}
\end{equation}

\noindent
To guide multi-scale extraction, $A_s$ is downsampled (stride=1 for $i=0$, stride=$4i$ for $i>0$), processed by a convolutional group $\mathcal{F}$ (Eq.~\ref{eq:F}), and upsampled.A residual connection yields the refined feature $A_m$. Finally, an attention map is generated and applied to $V_s$:
\begin{equation}
    H_i=\text{Sigmoid}(\text{Conv}_{1\times1}(A_m)) \odot V_s .
    \label{eq:Hi}
\end{equation}
Outputs from all $N$ heads are concatenated and fused. The subsequent FSG module enhances spatial-aware representation by splitting the feature into two parts $\{F_1,F_2\}$, generating a spatial weight mask from $F_1$, and applying it to $F_2$ for adaptive selection. After all MHCBs, a $1\times1$ convolution further refines features. The reconstruction module uses a $3\times3$ convolution and a PixelShuffle layer, aided by a global residual connection to recover high-frequency details.

%==========================================================
\noindent\textbf{Training Details. }The proposed AMCANet contains seven MHCBs, in which they set the number of feature maps to 32. Also, the head number of the AMCA is set to 2. Throughout the entire training process, they use the Adam optimizer, where $\beta{1}$ = 0.9 and $\beta{2}$ = 0.999. The model is trained for 1000k iterations in each stage. Input patches are randomly cropped and augmented. Data augmentation strategies included horizontal and vertical flips, and random rotations of 90, 180, and 270 degrees.
Model training was performed using Pytorch 1.12.0 on RTX 3090.
Specifically, the training strategy consists of several steps as follows.

1. In the starting stage, they train the model from scratch on the 800 images of DIV2K \cite{DIV2K} and the first 10k images of LSDIR \cite{lilsdir} datasets.
The model is trained for total $10^6$ iterations by minimizing L1 loss and FFT loss \cite{smfanet}. The HR patch size is set to 256$\times$256, while the mini-batch size is set to 64. They set the initial learning rate to 1 $\times$ $10^{-3}$ and the minimum one to 1 $\times$ $10^{-5}$, which is updated by the Cosine Annealing scheme.

2. In the second stage, they increase the HR patch size to 384, while the mini-batch size is set to 32. The model is fine-tuned by minimizing the L1 loss and FFT loss. They set the initial learning rate to 5 $\times$ $10^{-4}$ and the minimum one to 1 $\times$ $10^{-6}$, which is updated by the Cosine Annealing scheme.

% 3. In the third stage, the model is initialized with the pre-trained weights of Stage 2, and fine-tuned with larger HR patches of size 480x480. Other settings are the same as in the second stage.

3. In the last stage, the model is fine-tuned with 480×480 HR patches; however, the loss function is changed to minimize the combination of L2 loss and FFT loss. Other settings are the same as Stage 2.

%% file: teams/team18_DISP/main.tex
\subsection{DISP}
\paragraph{Method}
The network structure used in this method is inspired by TSSR \cite{ren2025tenth}. By introducing reparameterization and knowledge distillation techniques, this network can more effectively achieve high-quality image super-resolution reconstruction.
\paragraph{Training Details}
This method uses the DIV2K and LSDIR datasets for training. The overall training process can be divided into the following four steps:
\begin{enumerate}[leftmargin=*, label=\textbf{\arabic*.}]
 
 \item \textbf{Teacher Network Pre-training.} High-resolution image patches of size 256×256 are randomly cropped from the dataset. An L1 loss is used as the objective function to train a teacher network with a base of 64 channels. The training settings include a batch size of 32, an initial learning rate of $5\times10^{-5}$, and a cosine annealing learning rate scheduling strategy, for a total of 100 training epochs.

\item \textbf{Student Network Pre-training.} High-resolution image patches of size 256×256 are randomly cropped from the dataset. An L1 loss is used as the objective function to train a student network with a base of 32 channels. Training settings include a batch size of 32, an initial learning rate of $5\times10^{-5}$, and a cosine annealing learning rate scheduling strategy, for a total of 50 training epochs.

\item \textbf{Knowledge distillation training.} Knowledge distillation was performed on the student network using a combination of affinity loss \cite{He2020FAKD} and L1 loss. The initial learning rate was set to $1\times10^{-4}$, and a cosine annealing learning rate scheduling strategy was employed for 100 training epochs.

\item \textbf{Student network fine-tuning.} High-resolution image patches of size 512×512 were randomly cropped from the dataset, and the student network was fine-tuned with a batch size of 32 and a learning rate of $1\times10^{-5}$ to further improve model performance.
\end{enumerate}

%% file: teams/team20_JustTry/main.tex
\subsection{Just Try}

\noindent\textbf{Method Details:}
\begin{itemize}
    \item As a follow-up to our Enhanced Reparameterize Residual Network (ERRN) \cite{ren2024ninth}, we propose a lightweight variant named ERRN2 as shown in Fig.\ref{fig:errn2_overview}. The reparameterize block (RepB) contains four RepConvs, which consist of an ERB\cite{FMEN} and a multi-branch reparameterize block (MBRB). Both ERB and MBRB are reparameterization blocks, which use complex structures during the training phase and are converted to a 3×3 convolution during inference. Every RepB uses a Simple Attention (SA) module, similar to SPAN \cite{wan2024swift}, to enhance output features. In ERRN2, we use six RepBs, with the number of feature channels C set to 32.
    
\begin{figure}[!t]
        \centering
        \includegraphics[width=0.99\linewidth]{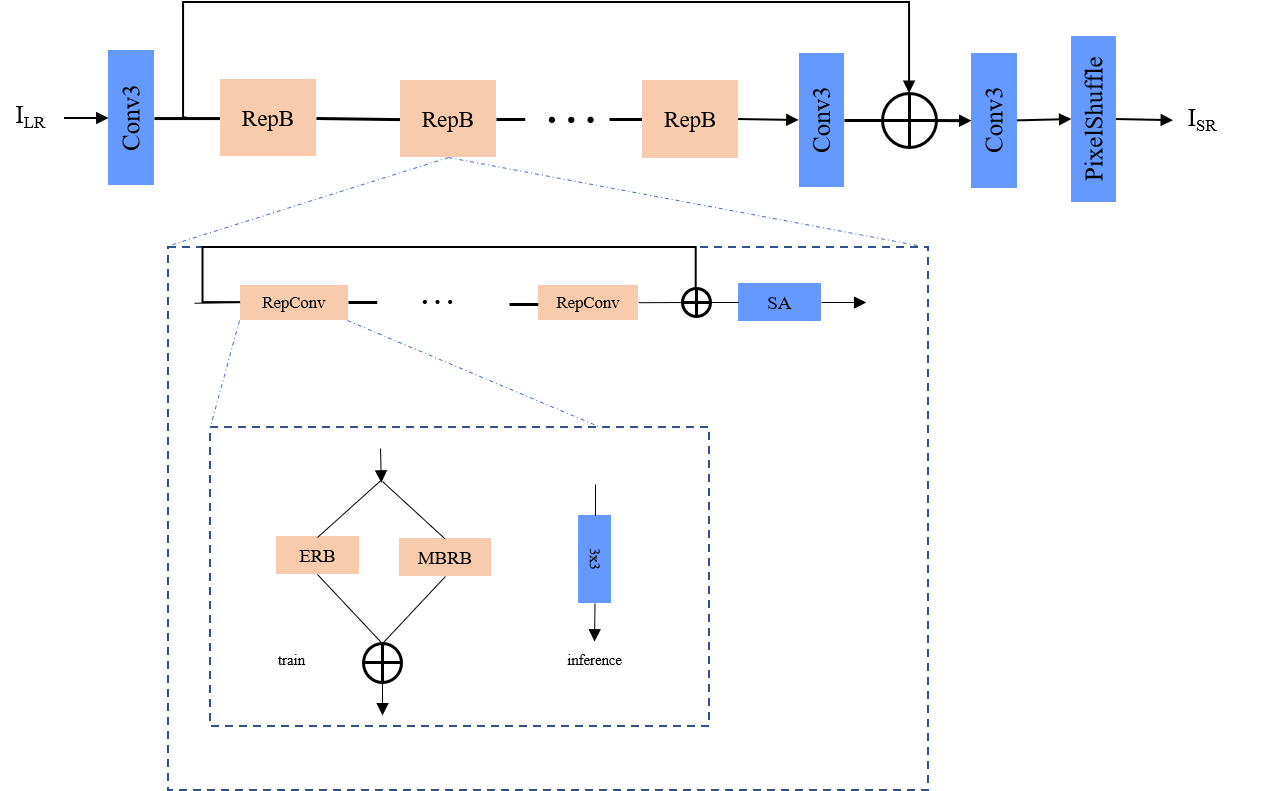}
        \caption{Team \textit{Just Try}: Overall structure of ERRN2}
        \label{fig:errn2_overview}
\end{figure}
    
    \item We use DIV2K, LSDIR, and Flickr2K datasets as training datasets. For each mini-batch, we randomly crop 16 patches from the LR images with the size of 64×64. We use cosine annealing learning scheme, the learning rate is initialized as $2\times10^{-4}$ and minimum learning rate is $1\times10^{-7}$, total 1000k iterations, the period of cosine is 250k iterations. We use Adam optimizer with $\beta_1=0.9$, $\beta_2=0.99$. The loss function is L1 loss. This training process is repeated five times, after which we increase the LR patch size to 128×128 and repeat the training three times.
    
    \item Then, we use L2 loss for fine-tuning on the same datasets, the LR size set to 128×128, initialize learning rate is $1\times10^{-4}$, total 1000k iterations, the period of cosine is 250k iterations. This fine-tuning is repeated three times, followed by increasing the LR size to 160×160 and repeating twice. All other settings remain the same as the previous stage.
\end{itemize}

%% file: teams/team21_MDAP/main.tex
\subsection{MDAP}
\label{ssec:team21}
\begin{figure*}[!t]
  \centering
  \includegraphics[width=\textwidth]{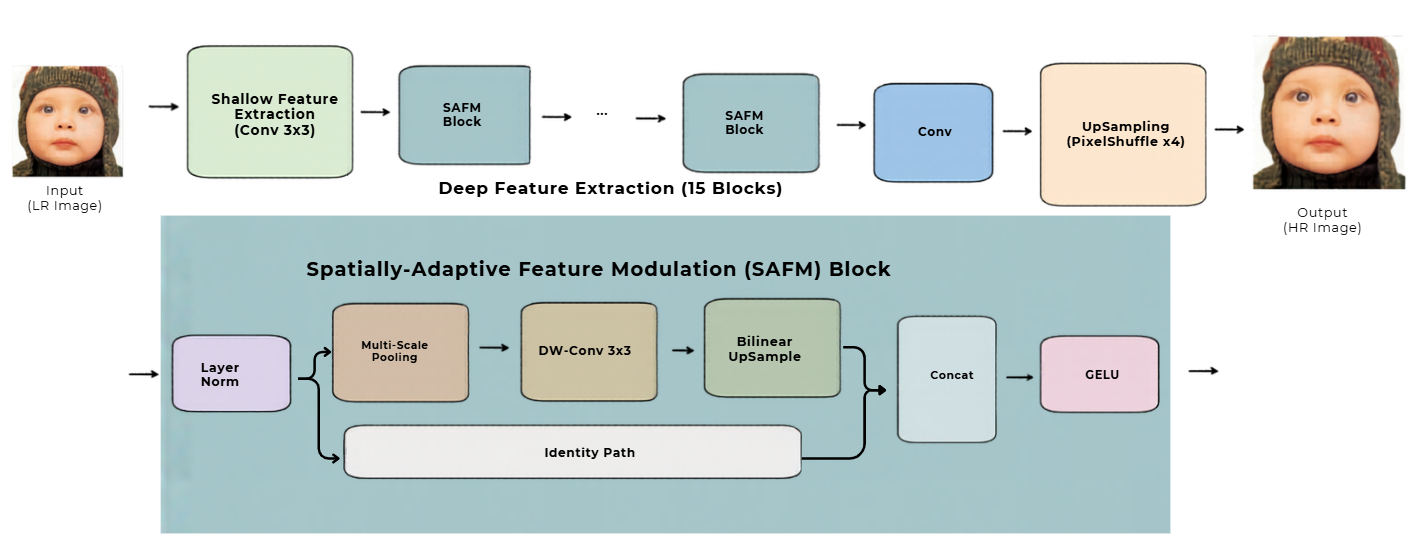}
  \caption{Team \textit{MDAP}: SAFMN-Deep15 overall architecture. The deep feature extraction path consists of 15 sequential SAFM blocks (utilizing multi-scale spatial pooling and GELU gating). The output is refined and added via a global residual before being finally upsampled by PixelShuffle($\times4$).}
  \label{fig:team21_mdap}
\end{figure*}

\textbf{Method Description.} MDAP presents \textbf{SAFMN-Deep15}, its submission to the NTIRE~2026 Efficient SR Challenge. The solution is built upon the Spatially-Adaptive Feature Modulation (SAFM) block~\cite{sun2023safmn}. 

SAFMN-Deep15 introduces a highly optimized architecture strictly adhering to the 0.15M parameter constraint while maximizing network depth. Inspired by prior work on multi-scale feature aggregation~\cite{sultan2025hybridatnet}, the model pushes the depth to 15 sequential SAFM blocks while keeping the feature dimension compact ($d=40$), prioritizing information processing depth over channel width.

\subsubsection{Model Architecture}
SAFMN-Deep15 is a pure-CNN model (0.149\,M parameters, 40 feature channels, 9.62 GFLOPs) comprising a shallow feature extraction layer, 15 sequential SAFM blocks, a global residual connection, and a PixelShuffle upsampling module.
The overall pipeline is shown in \cref{fig:team21_mdap} and proceeds as follows.

\begin{enumerate}[leftmargin=*, label=\textbf{\arabic*.}]
  \item \textbf{Shallow Feature Extraction.}
    A $3\times3$ convolution maps the input LR image to the 40-channel initial feature space.

  \item \textbf{Deep Feature Extraction.}
    Fifteen SAFM blocks are applied sequentially to capture long-range spatial context at multiple scales. Each SAFM block employs multi-scale pooling at resolutions $H, \frac{H}{2}, \frac{H}{4}, \frac{H}{8}$ and depthwise convolutions, yielding an element-wise GELU-gated attention map without the heavy computational cost of standard Transformer-based mechanisms. Each block also features a feed-forward network (FFN scale 2.0).

  \item \textbf{Feature Refinement and Global Residual.}
    The output of the 15 deep feature blocks is passed through a $3\times3$ refinement convolution (outputting 48 channels) and added element-wise to the bilinearly upsampled input image via a global residual skip connection, aiming to strictly preserve low-frequency content.

  \item \textbf{Reconstruction.}
    A PixelShuffle($\times4$) layer rearranges the 48-channel feature tensor into the final high-resolution RGB output.
\end{enumerate}

\subsubsection{Training Strategy}
The model leverages a progressive two-stage strategy combining structural fidelity and knowledge distillation.

\textbf{Stage~1: Standard Warm-Up.}
\begin{itemize}[leftmargin=*, topsep=2pt, itemsep=1pt]
  \item \textbf{Training data:} DIV2K (800 images) and LSDIR (84,991 images).
  \item \textbf{Patch size:} $64\!\times\!64$ LR crops.
  \item \textbf{Loss function:} Charbonnier Loss + Spatial Laplacian Frequency Loss ($\lambda = 0.05$).
  \item \textbf{Optimizer:} AdamW ($\beta_1=0.9$, $\beta_2=0.999$, weight decay $0.01$); initial LR $2\times10^{-4}$.
  \item \textbf{Batch size:} 32.
\end{itemize}

\textbf{Stage~2: Knowledge Distillation.}
\begin{itemize}[leftmargin=*, topsep=2pt, itemsep=1pt]
  \item \textbf{Teacher Network:} Pre-trained SPAN~\cite{wan2024swift} teacher model (\texttt{feature\_channels=28}).
  \item \textbf{Loss function:} MSE feature distillation ($\lambda_{\text{KD}} = 0.5$).
  \item \textbf{LR Schedule:} Decayed to $5\times10^{-5}$ using cosine annealing.
  \item \textbf{Early stopping:} Triggered at $\sim330$K iterations based on validation stability.
\end{itemize}

%% file: sec/5_appendix.tex
%-----------------------%
\appendix
%-----------------------%
\section{Teams and affiliations}
\label{sec:teams}
\subsection*{NTIRE 2025 team}
\noindent\textit{\textbf{Title: }} NTIRE 2025 Efficient Super-Resolution Challenge\\
\noindent\textit{\textbf{Members: }} \\
Bin Ren$^{1}$ (\href{mailto:bin.ren@mbzuai.ac.ae}{bin.ren@mbzuai.ac.ae}),\\
Hang Guo$^2$ (\href{cshguo@gmail.com }{cshguo@gmail.com}),\\
Yan Shu$^{3}$ (\href{mailto:bin.ren@unitn.it}{yan.shu@unitn.it}),\\
Jiaqi Ma$^{1}$ (\href{mailto:jiaqi.ma@mbzuai.ac.ae}{jiaqi.ma@mbzuai.ac.ae}),\\
Ziteng Cui$^{4}$ (\href{mailto:cui@mi.t.u-tokyo.ac.jp}{cui@mi.t.u-tokyo.ac.jp}),\\
Shuhong Liu$^{4}$ (\href{mailto:s-liu@mi.t.u-tokyo.ac.jp)}{s-liu@mi.t.u-tokyo.ac.jp)}),\\
Guofeng Mei$^{5}$ (\href{mailto:gmei@fbk.eu}{gmei@fbk.eu}),\\
Lei Sun$^{6}$ (\href{mailto:lei.sun@insait.ai}{lei.sun@insait.ai}),\\
Zongwei Wu$^7$ (\href{zongwei.wu@uni-wuerzburg.de}{zongwei.wu@uni-wuerzburg.de}),\\
Fahad Shahbaz Khan$^{1}$ (\href{mailto:fahad.khan@mbzuai.ac.ae}{fahad.khan@mbzuai.ac.ae}),\\
Salman Khan$^{1}$ (\href{mailto:salman.khan@mbzuai.ac.ae}{salman.khan@mbzuai.ac.ae}),\\
Radu Timofte$^{7}$ (\href{mailto:radu.timofte@uni-wuerzburg.de}{radu.timofte@vision.ee.ethz.ch})\\
Yawei Li$^8$ (\href{mailto:li.yawei.ai@gmail.com}{li.yawei.ai@gmail.com}),\\
\noindent\textit{\textbf{Affiliations: }}\\
$^1$ MBZUAI, UAE\\
$^2$ Tsinghua University, China\\
$^3$ University of Trento, Italy\\
$^4$ Univesity of Tokyo, Japan\\
$^5$ FBK, Italy\\
$^6$ INSAIT, Bulgaria\\
$^7$ University of W\"urzburg, Germany\\
$^8$ ETH Z\"urich, Switzerland\\

\input{teams/team22_XiaomiMM/affiliation}
\input{teams/team01_BOE_AIoT/affiliation}

\input{teams/team16_PKDSR/affiliation}

\input{teams/team04_ZenoSR/affiliation}

\input{teams/team05_CUIT_HTT/affiliation}

\input{teams/team06_HAESR/affiliation}

\input{teams/team09_IN2GM/affiliation}

\input{teams/team10_Sunflower/affiliation}

\input{teams/team11_XuptSR/affiliation}

\input{teams/team12_WMESR/affiliation}

\input{teams/team15_VARH-AI/affiliation}

\input{teams/team17_XSR/affiliation}

\input{teams/team18_DISP/affiliation}

\input{teams/team20_JustTry/affiliation}

\input{teams/team21_MDAP/affiliation}

%% file: teams/team22_XiaomiMM/affiliation.tex
 \subsection*{XiaomiMM}
  \noindent\textit{\textbf{Title: }} SPANV2: Beyond Parameter-free Attention for Efficient Super-Resolution\\
  \noindent\textit{\textbf{Members: }} \\
  Hongyuan Yu$^1$ (\href{mailto:yuhyuan1995@gmail.com}{yuhyuan1995@gmail.com}),\\
  Pufan Xu$^2$ (\href{mailto:xpf22@mails.tsinghua.edu.cn}{xpf22@mails.tsinghua.edu.cn}),\\
  Chen Wu$^3$ (\href{mailto:wuchen5@nudt.edu.cn}{wuchen5@nudt.edu.cn}),\\
  Long Peng$^4$ (\href{mailto:longp2001@mail.ustc.edu.cn}{longp2001@mail.ustc.edu.cn}),\\
  Jiaojiao Yi$^1$ (\href{mailto:yijiaojiao@xiaomi.com}{yijiaojiao@xiaomi.com}),\\
  Siyang Yi$^1$ (\href{mailto:yisiyang1@xiaomi.com}{yisiyang1@xiaomi.com}),\\
  Yuning Cui$^5$ (\href{mailto:yuning.cui@in.tum.de}{yuning.cui@in.tum.de}),\\
  Jingyuan Xia$^3$ (\href{mailto:j.xia10@nudt.edu.cn}{j.xia10@nudt.edu.cn}),\\
  Xing Mou$^3$ (\href{mailto:mouxing24@nudt.edu.cn}{mouxing24@nudt.edu.cn}),\\
  Keji He$^6$ (\href{mailto:keji.he@sdu.edu.cn}{keji.he@sdu.edu.cn}),\\
  Jinlin Wu$^7$ (\href{mailto:jinlin.wu@cair-cas.org.hk}{jinlin.wu@cair-cas.org.hk})\\
  \noindent\textit{\textbf{Affiliations: }} \\
  $^1$ Multimedia Department, Xiaomi Inc. \\
  $^2$ School of Integrated Circuits, Tsinghua University \\
  $^3$ NUDT \\
  $^4$ USTC \\
  $^5$ TU Munich \\
  $^6$ Shandong University \\
  $^7$ CAIR-CAS \\

%% file: teams/team01_BOE_AIoT/affiliation.tex
\subsection*{BOE\_AIoT}
\noindent\textit{\textbf{Title: }} A Pruned and Distilled Super-resolution Model\\
\noindent\textit{\textbf{Members: }} \\
Zongang Gao$^1$ (\href{mailto:yourEmail@xxx.xxx}{gaozongang@qq.com}),\\
Sen Yang$^1$, \\
Rui Zheng$^1$, \\
Fengguo Li$^1$, \\
\noindent\textit{\textbf{Affiliations: }} \\ 
$^1$ AIoT CTO, BOE Technology Group Co.,Ltd.  \\

%% file: teams/team16_PKDSR/affiliation.tex
\subsection*{PKDSR}
\noindent\textit{\textbf{Title: }} \textbf{P}runing with \textbf{K}nowledge \textbf{D}istillation for \textbf{S}uper-\textbf{R}esolution\\
\noindent\textit{\textbf{Members: }} \\
Yecheng Lei$^1$ (\href{yechenglei@smail.nju.edu.cn}{yechenglei@smail.nju.edu.cn}),\\
Wenkai Min$^2$(\href{241870230@smail.nju.edu.cn}{241870230@smail.nju.edu.cn}), \\
Jie Liu$^3$(\href{liujie@nju.edu.cn}{liujie@nju.edu.cn}) \\
\noindent\textit{\textbf{Affiliations: }} \\ 
$^1$ Nanjing University

%% file: teams/team04_ZenoSR/affiliation.tex
\subsection*{ZenoSR}
\noindent\textit{\textbf{Title: }} Adaptive Calibration Network for Efficient Image Super-Resolution\\
\noindent\textit{\textbf{Members: }} \\
Qingliang Liu$^1$ (\href{mailto:yourEmail@xxx.xxx}{liuqingliang1@honor.com}),\\
Yang Cheng$^2$ (\href{mailto:yourEmail@xxx.xxx}{yangcheng24@m.fudan.edu.cn})\\
\noindent\textit{\textbf{Affiliations: }} \\ 
$^1$ Honor Device Co.,Ltd. \\
$^2$ State Key Laboratory of Integrated Chip and System, Fudan University \\

%% file: teams/team05_CUIT_HTT/affiliation.tex
\subsection*{CUIT\_HTT}

\noindent\textit{\textbf{Title: }}SPAN-Mamba: Revisiting Lightweight Super-Resolution with State-Space Gating and Distillation\\
\noindent\textit{\textbf{Members: }} \\
Jing Hu$^1$ (\href{mailto:jing_hu@163.com}{jing\_hu@163.com}),\\
Xuan Zhang$^1$, \\
Rui Ding$^1$, \\
Tingyi Zhang$^1$, \\
Hui Deng$^1$, \\
\noindent\textit{\textbf{Affiliations: }} \\ 
$^1$ Chengdu University of Information Technology

%% file: teams/team06_HAESR/affiliation.tex
\subsection*{HAESR}
\noindent\textit{\textbf{Title: }} Efficient Image Super-Resolution via Hybrid Attention and Attention Sharing\\
\noindent\textit{\textbf{Members: }} \\
Mingyang Li$^1$ (\href{mailto:2024222055110@stu.scu.edu.cn}{2024222055110@stu.scu.edu.cn}),\\
Guanglu Dong$^1$, \\
Zheng Yang$^1$, \\
Chao Ren$^1$, \\
\noindent\textit{\textbf{Affiliations: }} \\ 
$^1$ Sichuan University \\

%% file: teams/team09_IN2GM/affiliation.tex
\subsection*{IN2GM}
\noindent\textit{\textbf{Title: }} RFDN-SPAN: Improving Efficiency with Parameter-Free Attention\\
\noindent\textit{\textbf{Members: }} \\
Matin Fazel$^1$ (\href{matin.fazel@concordia.ca}{matin.fazel@concordia.ca}),\\
Abdelhak Bentaleb$^1$ (\href{abdelhak.bentaleb@concordia.ca}{abdelhak.bentaleb@concordia.ca}),\\
\noindent\textit{\textbf{Affiliations: }} \\ 
$^1$ Concordia University \\

%% file: teams/team10_Sunflower/affiliation.tex
\subsection*{Sunflower}
\noindent\textit{\textbf{Title: }} Hierarchical Feature Enhancement Network for Efficient Single Image Super-Resolution\\
\noindent\textit{\textbf{Members: }} \\
Zitao Dai 1$^1$ (\href{1709358028@qq.com}{1709358028@qq.com}),\\
\noindent\textit{\textbf{Affiliations: }} \\ 
$^1$ Southwest Jiaotong University \\

%% file: teams/team11_XuptSR/affiliation.tex
\subsection*{XuptSR}
\noindent\textit{\textbf{Title: }} Variance-Guided Spatial-Channel Context Interaction Network for Efficient
Image Super-Resolution\\
\noindent\textit{\textbf{Members: }} \\
Mengyang Wang$^1$ (\href{mailto:yourEmail@xxx.xxx}{yang813@stu.xupt.edu.cn}),\\
Fulin Liu$^1$, \\
Jing Wei$^1$, \\
Qian Wang$^{1,2}$, \\
Hongying Liu$^{3}$ \\
\noindent\textit{\textbf{Affiliations: }} \\ 
$^1$ Xi'an University of Posts and Telecommunications, Xi'an, China \\
$^2$ National Engineering Laboratory for Cyber Event Warning and Control Technologies \\
$^3$ Medical School of Tianjin University \\

%% file: teams/team12_WMESR/affiliation.tex
\subsection*{WMESR}
\noindent\textit{\textbf{Title: }} DualWave-Mamba: A Wavelet-Guided Dual-Branch Network for Efficient Image Super-Resolution\\
\noindent\textit{\textbf{Members: }} \\
Hongbo Fang$^1$ (\href{3129457147@nuaa.edu.cn}{3129457147@nuaa.edu.cn}),\\
Lingxuan Li$^1$ (\href{2540526357@qq.com}
{2540526357@qq.com}), \\
Lin Si$^1$ (\href{sgin.16_23@nuaa.edu.cn}
{sgin.16\_23@nuaa.edu.cn}), \\
Pan Gao$^1$ (\href{Pan.Gao@nuaa.edu.cn}
{Pan.Gao@nuaa.edu.cn}), \\
Moncef Gabbouj$^2$ (\href{Moncef.Gabbouj@tuni.fi}
{Moncef.Gabbouj@tuni.fi}), \\
\noindent\textit{\textbf{Affiliations: }} \\ 
$^1$ Nanjing University of Aeronautics and Astronautics \\
$^2$ Tampere University \\

%% file: teams/team15_VARH-AI/affiliation.tex
\subsection*{VARH-AI}
\noindent\textit{\textbf{Title: }} Zero-Cost Operator Fusion with DSCLoRA Distillation and MuAdamW Optimization\\
\noindent\textit{\textbf{Members: }} \\
Shubham Sharma$^1$ (\href{p20240036@pilani.bits-pilani.ac.in}{p20240036@pilani.bits-pilani.ac.in}),\\
Manish Prasad$^1$ (\href{p20240903@pilani.bits-pilani.ac.in}{p20240903@pilani.bits-pilani.ac.in})\\
\noindent\textit{\textbf{Affiliations: }} \\ 
$^1$ Birla Institute of Technology Pilani, Pilani Campus, Rajasthan 333031, India \\

%% file: teams/team17_XSR/affiliation.tex
\subsection*{XSR}
\noindent\textit{\textbf{Title: }} AMCANet: A Lightweight Architecture-guided Multi-head Convolution Attention Network for Efficient Image Super-Resolution\\
\noindent\textit{\textbf{Members: }} \\
Rui Chen$^1$ (\href{mailto:chenr269@163.com}{chenr269@163.com}),\\
Shurui Shi$^1$,\\
\noindent\textit{\textbf{Affiliations: }} \\ 
$^1$ Shenzhen International Graduate School, Tsinghua University, China \\

%% file: teams/team18_DISP/affiliation.tex
\subsection*{DISP}
\noindent\textit{\textbf{Title: }} Affinity Matrix Distillation for Lightweight Image Super-Resolution\\
\noindent\textit{\textbf{Members: }} \\
Keye Cao$^1$ (\href{mailto:clickcao@outlook.com}{clickcao@outlook.com}),\\

\noindent\textit{\textbf{Affiliations: }} \\

%% file: teams/team20_JustTry/affiliation.tex
\subsection*{JustTry}
\noindent\textit{\textbf{Title: }} Lightweight Enhanced Reparameterize Residual Network for Efficient Image
Super-Resolution\\
\noindent\textit{\textbf{Members: }} \\
Haobo Li $^1$ (\href{mailto:yourEmail@xxx.xxx}{1114940412@qq.com})

\noindent\textit{\textbf{Affiliations: }} \\

%% file: teams/team21_MDAP/affiliation.tex
\subsection*{MDAP}
\noindent\textit{\textbf{Title: }} SAFMN-Deep15: Spatially-Adaptive Feature Modulation Network with Deep 15-Block Architecture\\
\noindent\textit{\textbf{Members: }} \\
Watchara Ruangsang$^1$ (\href{mailto:watchara.ruan@kmutt.ac.th}{watchara.ruan@kmutt.ac.th}),\\
Supavadee Aramvith$^2$ (\href{mailto:supavadee.a@chula.ac.th}{supavadee.a@chula.ac.th})\\
\noindent\textit{\textbf{Affiliations: }} \\ 
$^1$ Media Technology Program, King Mongkut's University of Technology Thonburi, Bangkok 10150, Thailand \\
$^2$ Multimedia Data Analytics and Processing Research Unit, Department of Electrical Engineering, Faculty of Engineering, Chulalongkorn University, Bangkok 10330, Thailand \\

%% file: main.bib
@String(CVPR= {IEEE Conf. Comput. Vis. Pattern Recog.})

@String(ICCV= {Int. Conf. Comput. Vis.})

@String(ECCV= {Eur. Conf. Comput. Vis.})

@String(TIP  = {IEEE Trans. Image Process.})

@String(TMM  = {IEEE Trans. Multimedia})

@String(ICASSP=	{ICASSP})

@String(ICIP = {IEEE Int. Conf. Image Process.})

@String(ICLR = {Int. Conf. Learn. Represent.})

@String(PR   = {PR})

@String(AAAI = {AAAI})

@String(CVPR  = {CVPR})

@String(ICCV  = {ICCV})

@String(ECCV  = {ECCV})

@String(TIP   = {IEEE TIP})

@String(TCSVT = {IEEE TCSVT})

@String(TMM   =	{IEEE TMM})

@String(ICIP  = {ICIP})

@String(ICLR  = {ICLR})

@String(PR = {PR})

@inproceedings{lilsdir,
  title={LSDIR: A Large Scale Dataset for Image Restoration},
  author={Li, Yawei and Zhang, Kai and Liang, Jingyun and Cao, Jiezhang and Liu, Ce and Gong, Rui and Zhang, Yulun and Tang, Hao and Liu, Yun and Demandolx, Denis and others},
  booktitle={CVPR Workshops},
  year={2023}
}

@inproceedings{flickr2k,
  title={Ntire 2017 challenge on single image super-resolution: Methods and results},
  author={Timofte, Radu and Agustsson, Eirikur and Van Gool, Luc and Yang, Ming-Hsuan and Zhang, Lei},
  booktitle={CVPR workshops},
  pages={114--125},
  year={2017}
}

@article{mamba,
  title={Mamba: Linear-time sequence modeling with selective state spaces},
  author={Gu, Albert and Dao, Tri},
  journal={arXiv preprint arXiv:2312.00752},
  year={2023}
}

@inproceedings{IMDN,
  title={Lightweight image super-resolution with information multi-distillation network},
  author={Hui, Zheng and Gao, Xinbo and Yang, Yunchu and Wang, Xiumei},
  booktitle={ACM MM},
  pages={2024--2032},
  year={2019}
}

@inproceedings{DIV2K,
    author={Timofte, Radu and Agustsson, Eirikur and Van Gool, Luc and Yang, Ming-Hsuan and Zhang, Lei and others},
    title={{NTIRE} 2017 Challenge on Single Image Super-Resolution: Methods and Results},
    booktitle={CVPR Workshops},
    year= {2017}
}

@inproceedings{agustsson2017ntire,
  title={{NTIRE} 2017 challenge on single image super-resolution: Dataset and study},
  author={Agustsson, Eirikur and Timofte, Radu},
  booktitle={CVPR Workshops},
  pages={126--135},
  year={2017}
}

@inproceedings{wan2024swift,
  title={Swift parameter-free attention network for efficient super-resolution},
  author={Wan, Cheng and Yu, Hongyuan and Li, Zhiqi and Chen, Yihang and Zou, Yajun and Liu, Yuqing and Yin, Xuanwu and Zuo, Kunlong},
  booktitle={CVPR},
  pages={6246--6256},
  year={2024}
}

@inproceedings{ren2025tenth,
  title={The tenth NTIRE 2025 efficient super-resolution challenge report},
  author={Ren, Bin and Guo, Hang and Sun, Lei and Wu, Zongwei and Timofte, Radu and Li, Yawei and others},
  booktitle={CVPR},
  pages={917--966},
  year={2025}
}

@inproceedings{FMEN,
  title={Fast and memory-efficient network towards efficient image super-resolution},
  author={Du, Zongcai and Liu, Ding and Liu, Jie and Tang, Jie and Wu, Gangshan and Fu, Lean},
  booktitle={CVPR},
  pages={853--862},
  year={2022}
}

@inproceedings{miao2025langhops,
  title={LangHOPS: Language Grounded Hierarchical Open-Vocabulary Part Segmentation},
  author={Miao, Yang and Zaech, Jan-Nico and Wang, Xi and Despinoy, Fabien and Pani Paudel, Danda and Van Gool, Luc},
  booktitle={NeurIPS},
  year={2025}
}

@misc{liu2020residualfeaturedistillationnetwork,
      title={Residual Feature Distillation Network for Lightweight Image Super-Resolution}, 
      author={Jie Liu and Jie Tang and Gangshan Wu},
      year={2020},
      eprint={2009.11551},
      archivePrefix={arXiv},
      primaryClass={eess.IV},
      url={https://arxiv.org/abs/2009.11551}, 
}

@inproceedings{guo2025mambav2,
  title={MambaIRv2: Attentive State Space Restoration},
  author={Guo, Hang and Guo, Yong and Zha, Yaohua and Zhang, Yulun and Li, Wenbo and Dai, Tao and Xia, Shu-Tao and Li, Yawei},
  booktitle={CVPR},
  year={2025}
}

@inproceedings{smfanet,
    title={SMFANet: A Lightweight Self-Modulation Feature Aggregation Network for Efficient Image Super-Resolution},
    author={Zheng, Mingjun and Sun, Long and Dong, Jiangxin and Pan, Jinshan},
    booktitle={ECCV},
    year={2024}
 }

@inproceedings{zamir2020mirnet,
    title={Learning Enriched Features for Real Image Restoration and Enhancement},
    author={Syed Waqas Zamir and Aditya Arora and Salman Khan and Munawar Hayat
            and Fahad Shahbaz Khan and Ming-Hsuan Yang and Ling Shao},
    booktitle={ECCV},
    year={2020}
}

@inproceedings{park2025efficient,
  title={Efficient attention-sharing information distillation transformer for lightweight single image super-resolution},
  author={Park, Karam and Soh, Jae Woong and Cho, Nam Ik},
  booktitle={AAAI},
  volume={39},
  pages={6416--6424},
  year={2025}
}

@article{hu2025large,
  title={Large Kernel Modulation Network for Efficient Image Super-Resolution},
  author={Hu, Quanwei and Tang, Yinggan and Zhang, Xuguang},
  journal={arXiv preprint arXiv:2508.11893},
  year={2025}
}

@inproceedings{wang2023omni,
  title={Omni aggregation networks for lightweight image super-resolution},
  author={Wang, Hang and Chen, Xuanhong and Ni, Bingbing and Liu, Yutian and Liu, Jinfan},
  booktitle={CVPR},
  pages={22378--22387},
  year={2023}
}

@inproceedings{chen2023activating,
  title={Activating more pixels in image super-resolution transformer},
  author={Chen, Xiangyu and Wang, Xintao and Zhou, Jiantao and Qiao, Yu and Dong, Chao},
  booktitle={CVPR},
  pages={22367--22377},
  year={2023}
}

@inproceedings{SwinIR,
  title={SwinIR: Image Restoration Using Swin Transformer}, 
  author={Jingyun Liang and Jiezhang Cao and Guolei Sun and Kai Zhang and Luc Van Gool and Radu Timofte},
  year={2021},
  booktitle={ICCVW}
}

@inproceedings{wang2025aim,
  title={AIM 2025 challenge on inverse tone mapping report: Methods and results},
  author={Wang, Chao and Banterle, Francesco and Ren, Bin and Timofte, Radu and Lu, Xin and Peng, Yufeng and Ge, Chengjie and Sun, Zhijing and Zhou, Ziang and Li, Zihao and others},
  booktitle={ICCVW},
  pages={5571--5584},
  year={2025}
}

@misc{SPAN,
  title={Swift Parameter-free Attention Network for Efficient Super-Resolution}, 
  author={Cheng Wan and Hongyuan Yu and Zhiqi Li and Yihang Chen and Yajun Zou and Yuqing Liu and Xuanwu Yin and Kunlong Zuo},
  year={2024},
  eprint={2311.12770},
  archivePrefix={arXiv},
  primaryClass={eess.IV},
  url={https://arxiv.org/abs/2311.12770}, 
}

@inproceedings{zhao2024efficient,
  title={Efficient single image super-resolution with entropy attention and receptive field augmentation},
  author={Zhao, Xiaole and Li, Linze and Xie, Chengxing and Zhang, Xiaoming and Jiang, Ting and Lin, Wenjie and Liu, Shuaicheng and Li, Tianrui},
  booktitle={ACM MM},
  pages={1302--1310},
  year={2024}
}

@inproceedings{li2023lsknet,
  title={Lsknet: Large selective kernel network for remote sensing object detection},
  author={Li, Yuxuan and Hou, Qibin and Zheng, Z},
  booktitle={ICCV},
  pages={4--6},
  year={2023}
}

@inproceedings{He2020FAKD,
  title={Fakd: Feature-Affinity Based Knowledge Distillation for Efficient Image Super-Resolution},
  author={He, Zibin and Dai, Tao and Lu, Jian and Jiang, Yong and Xia, Shu-Tao},
  booktitle={ICIP},
  pages={518--522},
  year={2020},
  organization={IEEE},
  doi={10.1109/ICIP40778.2020.9190917}
}

@article{chai2025dsclora,
  author       = {Xinning Chai and
                  Yao Zhang and
                  Yuxuan Zhang and
                  Zhengxue Cheng and
                  Yingsheng Qin and
                  Yucai Yang and
                  Li Song},
  title        = {Distillation-Supervised Convolutional Low-Rank Adaptation for Efficient
                  Image Super-Resolution},
  journal      = {CoRR},
  volume       = {abs/2504.11271},
  year         = {2025},
  url          = {https://doi.org/10.48550/arXiv.2504.11271},
  doi          = {10.48550/ARXIV.2504.11271},
  eprinttype   = {arXiv},
  eprint       = {2504.11271},
  bibsource    = {dblp computer science bibliography, https://dblp.org}
}

@misc{ultralytics2026yolo26,
  title={Ultralytics {YOLO26} Documentation},
  author={{Ultralytics}},
  howpublished={\url{https://docs.ultralytics.com/models/yolo26/}},
  year={2026},
  url={https://docs.ultralytics.com/models/yolo26/}
}

@inproceedings{ren2024ninth,
  title={The ninth NTIRE 2024 efficient super-resolution challenge report},
  author={Ren, Bin and Li, Yawei and Mehta, Nancy and Timofte, Radu and Yu, Hongyuan and Wan, Cheng and Hong, Yuxin and Han, Bingnan and Wu, Zhuoyuan and Zou, Yajun and others},
  booktitle={CVPR},
  pages={6595--6631},
  year={2024}
}

@inproceedings{ntire26deepfake, 
title={{    Robust Deepfake Detection, NTIRE 2026 Challenge: Report    }}, 
author={    Hopf, Benedikt and  Timofte, Radu and others    },   
booktitle={CVPR (CVPR) Workshops},  
year = {2026} 
}

@inproceedings{ntire26hrdepth, 
title={{    NTIRE 2026 Challenge on High-Resolution Depth of non-Lambertian Surfaces    }}, 
author={    Zama Ramirez, Pierluigi and  Tosi, Fabio and  Di Stefano, Luigi and  Timofte, Radu and  Costanzino, Alex and  Poggi, Matteo and  Salti, Samuele and  Mattoccia, Stefano and others    },   
booktitle={CVPR (CVPR) Workshops},  
year = {2026} 
}

@inproceedings{ntire26raim_fusion, 
title={{    NTIRE 2026 The 3rd Restore Any Image Model (RAIM) Challenge: Multi-Exposure Image Fusion in Dynamic Scenes (Track2)    }}, 
author={    Qu, Lishen and  Liu, Yao and  Liang, Jie and  Zeng, Hui and  Dai, Wen and  Guan, Ya-nan and  Qin, Guanyi and  Zhou, Shihao and  Yang, Jufeng and  Zhang, Lei and  Timofte, Radu and others    },   
booktitle={CVPR (CVPR) Workshops},  
year = {2026} 
}

@inproceedings{ntire26raim_portrait, 
title={{NTIRE 2026 The 3rd Restore Any Image Model (RAIM) Challenge: AI Flash Portrait (Track 3)    }}, 
author={Guan, Ya-nan and  Zhang, Shaonan and  Guo, Hang and  Wang, Yawen and  Fan, Xinying and  Liang, Jie and  Zeng, Hui and  Qin, Guanyi and  Qu, Lishen and  Dai, Tao and  Xia, Shu-Tao and  Zhang, Lei and  Timofte, Radu and others    },   
booktitle={CVPR (CVPR) Workshops},  
year = {2026} 
}

@inproceedings{ntire26raim_piqa, 
title={{NTIRE 2026 The 3rd Restore Any Image Model (RAIM) Challenge: Professional Image Quality Assessment (Track 1)    }}, 
author={Qin, Guanyi and  Liang, Jie and  Zhang, Bingbing and  Qu, Lishen and  Guan, Ya-nan and  Zeng, Hui and  Zhang, Lei and  Timofte, Radu and others    },   
booktitle={CVPR (CVPR) Workshops},  
year = {2026} 
}

@inproceedings{ntire26lightsr, 
title={{NTIRE 2026 Challenge on Light Field Image Super-Resolution: Methods and Results    }}, 
author={Wang, Yingqian and  Liang, Zhengyu and  Zhang, Fengyuan and  Zhao, Wending and  Wang, Longguang and  Li, Juncheng and  Yang, Jungang and  Timofte, Radu and  Guo, Yulan and others    },   
booktitle={CVPR (CVPR) Workshops},  
year = {2026} 
}

@inproceedings{ntire263dsr, 
title={{NTIRE 2026 Challenge on 3D Content Super-Resolution: Methods and Results    }}, 
author={Wang, Longguang and  Guo, Yulan and  Wang, Yingqian and  Li, Juncheng and  Peng, Sida and  Zhang, Ye and  Timofte, Radu and  Chen, Minglin and  Wang, Yi and  Hu, Qibin and  Lei, Wenjie and others    },   
booktitle={CVPR (CVPR) Workshops},  
year = {2026} 
}

@inproceedings{ntire26videores, 
title={{NTIRE 2026 Challenge on Bitstream-Corrupted Video Restoration: Methods and Results    }}, 
author={Zou, Wenbin and  Liu, Tianyi and  Wu, Kejun and  Zhuang, Huiping and  Wu, Zongwei and  Zhou, Zhuyun and  Timofte, Radu and  others     },   booktitle={CVPR (CVPR) Workshops},  
year = {2026} 
}

@inproceedings{ntire26XAIGCqa, 
title={{NTIRE 2026 X-AIGC Quality Assessment Challenge: Methods and Results    }}, 
author={Liu, Xiaohong and  Min, Xiongkuo and  Zhai, Guangtao and  Hu, Qiang and  Cao, Jiezhang and  Zhou, Yu and  Sun, Wei and  Wen, Farong and  Xu, Zitong and  Zhou, Yingjie and  Duan, Huiyu and  Liu, Lu and  Wang, Jiarui and  Luo, Siqi and  Li, Chunyi and  Xu, Li and  Zhang, Zicheng and  Shi, Yue and  Wang, Yubo and  Zhang, Minghong and  Guo, Chunchao and  Hu, Zhichao and  Chen, Mingtao and  Wu, Xiele and  Ma, Xin and  Lv, Zhaohe and  Xue, Yuanhao and  Wang, Jiaqi and  Sha, Xinxing and  Timofte, Radu and  others},   
booktitle={CVPR (CVPR) Workshops},  
year = {2026} 
}

@inproceedings{ntire26shadow, 
title={{Advances in Single-Image Shadow Removal: Results from the NTIRE 2026 Challenge}}, 
author={Vasluianu, Florin-Alexandru and  Seizinger, Tim and  Zhou, Zhuyun and  Wu, Zongwei and  Timofte, Radu and  others},   
booktitle={CVPR (CVPR) Workshops},  
year = {2026} 
}

@inproceedings{ntire26lightnorm, 
title={{Learning-Based Ambient Lighting Normalization: NTIRE 2026 Challenge Results and Findings}}, 
author={Vasluianu, Florin-Alexandru and  Seizinger, Tim and  Chen, Jeffrey and  Zhou, Zhuyun and  Wu, Zongwei and  Timofte, Radu and  others    },   booktitle={CVPR (CVPR) Workshops},  
year = {2026} 
}

@inproceedings{ntire26bokeh, 
title={{The First Controllable Bokeh Rendering Challenge at NTIRE 2026}}, 
author={Seizinger, Tim and  Vasluianu, Florin-Alexandru and  Conde, Marcos V. and  Chen, Jeffrey and  Zhou, Zhuyun and  Wu, Zongwei and  Timofte, Radu and  others    },   
booktitle={CVPR (CVPR) Workshops},  
year = {2026} 
}

@inproceedings{ntire26ripdetseg, 
title={{NTIRE 2026 Rip Current Detection and Segmentation (RipDetSeg) Challenge Report}}, 
author={Dumitriu, Andrei and  Ralhan, Aakash and  Miron, Florin and  Tatui, Florin and  Ionescu, Radu Tudor and  Timofte, Radu and  others},   booktitle={CVPR (CVPR) Workshops},  
year = {2026} 
}

@inproceedings{ntire26llie, 
title={{Low Light Image Enhancement Challenge at NTIRE 2026 }}, 
author={Ciubotariu, George and  S M A,  Sharif and  Rehman, Abdur and  Ali, Fayaz and  Naqvi, Rizwan Ali and  Conde, Marcos and  Timofte, Radu and others},   
booktitle={CVPR (CVPR) Workshops},  
year = {2026} 
}

@inproceedings{ntire26highfps, 
title={{High FPS Video Frame Interpolation Challenge at NTIRE 2026}}, 
author={Ciubotariu, George and  Zhou, Zhuyun and  Jin, Yeying and  Wu, Zongwei and  Timofte, Radu and  others    },   
booktitle={CVPR (CVPR) Workshops},  
year = {2026} 
}

@inproceedings{ntire26nthaze, 
title={{NT-HAZE: A Benchmark Dataset for Realistic Night-time Image Dehazing}}, 
author={Ancuti, Radu and  Ancuti, Codruta and  Timofte, Radu and  Ancuti, Cosmin    },   
booktitle={CVPR (CVPR) Workshops},  
year = {2026} 
}

@inproceedings{ntire26nthaze_rep, 
title={{NTIRE 2026 Nighttime Image Dehazing Challenge Report}}, 
author={Ancuti, Radu and  Brateanu, Alexandru and  Vasluianu, Florin and  Balmez, Raul and  Orhei, Ciprian and  Ancuti, Codruta and  Timofte, Radu and  Ancuti, Cosmin and others},   
booktitle={CVPR (CVPR) Workshops},  
year = {2026} 
}

@inproceedings{ntire26isp, 
title={{NTIRE 2026 Challenge on Learned Smartphone ISP with Unpaired Data: Methods and Results}}, 
author={Perevozchikov, Georgy and  Vladimirov, Daniil and  Timofte, Radu and  others},   
booktitle={CVPR (CVPR) Workshops},  
year = {2026} 
}

@inproceedings{ntire26ugcvideo, 
title={{NTIRE 2026 Challenge on Short-form UGC Video Restoration in the Wild with Generative Models: Datasets, Methods and Results    }}, author={    Li, Xin and  Gong, Jiachao and  Wang, Xijun and  Xiong, Shiyao and  Li, Bingchen and  Yao, Suhang  and  Zhou, Chao and  Chen, Zhibo and  Timofte, Radu and others    },   
booktitle={CVPR (CVPR) Workshops},  
year = {2026} 
}

@inproceedings{ntire26dual_focus, 
title={{NTIRE 2026 The Second Challenge on Day and Night Raindrop Removal for Dual-Focused Images: Methods and Results    }}, 
author={Li, Xin and  Jin, Yeying and  Yao, Suhang and  Lin, Beibei and  Fan, Zhaoxin and   Yan, Wending and  Jin, Xin and  Wu, Zongwei  and  Li, Bingchen  and  Shi, Peishu and  Yang, Yufei and  Li, Yu and  Chen, Zhibo  and  Wen, Bihan and  Tan, Robby and  Timofte, Radu and others},   
booktitle={CVPR (CVPR) Workshops},  
year = {2026} 
}

@inproceedings{ntire26srx4, 
title={{The Fourth Challenge on Image Super-Resolution (×4) at NTIRE 2026: Benchmark Results and Method Overview    }}, 
author={Chen, Zheng and  Liu, Kai and  Wang, Jingkai and  Yan, Xianglong and  Li, Jianze and  Zhang, Ziqing and  Gong, Jue and  Li, Jiatong and  Sun, Lei and  Liu, Xiaoyang and  Timofte, Radu and  Zhang, Yulun and others    },   
booktitle={CVPR (CVPR) Workshops},  
year = {2026} 
}

@inproceedings{ntire26retouching, 
title={{Photography Retouching Transfer, NTIRE 2026 Challenge: Report    }}, 
author={Elezabi, Omar and  V. Conde, Marcos and  Wu, Zongwei and  Jin, Yeying and  Timofte, Radu and others    },   
booktitle={CVPR (CVPR) Workshops},  
year = {2026} 
}

@inproceedings{ntire26rwsr, 
title={{The First Challenge on Mobile Real-World Image Super-Resolution at NTIRE 2026: Benchmark Results and Method Overview}}, 
author={Li, Jiatong and  Chen, Zheng and  Liu, Kai and  Wang, Jingkai and  Zhou, Zihan and  Liu, Xiaoyang and  Zhu, Libo and  Timofte, Radu and  Zhang, Yulun and others    },   
booktitle={CVPR (CVPR) Workshops},  
year = {2026} 
}

@inproceedings{ntire26rsirsr, 
title={{The First Challenge on Remote Sensing Infrared Image Super-Resolution at NTIRE 2026: Benchmark Results and Method Overview}}, author={    Liu, Kai and  Yue, Haoyang and  Lin, Zeli and  Chen, Zheng and  Wang, Jingkai and  Gong, Jue and  Timofte, Radu and  Zhang, Yulun and  others    },   
booktitle={CVPR (CVPR) Workshops},  
year = {2026} 
}

@article{an2025onestory,
  title={OneStory: Coherent Multi-Shot Video Generation with Adaptive Memory},
  author={An, Zhaochong and Jia, Menglin and Qiu, Haonan and Zhou, Zijian and Huang, Xiaoke and Liu, Zhiheng and Ren, Weiming and Kahatapitiya, Kumara and Liu, Ding and He, Sen and others},
  journal={CVPR},
  year={2026}
}

@inproceedings{ntire26aigendet, 
title={{    NTIRE 2026 Challenge on Robust AI-Generated Image Detection in the Wild    }}, 
author={    Gushchin, Aleksandr and  Abud, Khaled and  Shumitskaya, Ekaterina and  Filippov, Artem and  Bychkov, Georgii and  Lavrushkin, Sergey and  Erofeev, Mikhail and  Antsiferova, Anastasia and  Chen, Changsheng and  Tan, Shunquan and  Timofte, Radu and  Vatolin, Dmitriy and others    },
booktitle={CVPR (CVPR) Workshops},  
year = {2026} 
}

@inproceedings{ntire26cdfsod, 
title={{The Second Challenge on Cross-Domain Few-Shot Object Detection at NTIRE 2026: Methods and Results}}, 
author={Qiu, Xingyu and  Fu, Yuqian and  Geng, Jiawei and  Ren, Bin and  Pan, Jiancheng and  Wu, Zongwei and  Tang, Hao and  Fu, Yanwei and  Timofte, Radu and  Sebe, Nicu and  Elhoseiny, Mohamed and others    },   
booktitle={CVPR (CVPR) Workshops},  
year = {2026} 
}

@inproceedings{ntire26finrec, 
title={{    NTIRE 2026 Challenge on End-to-End Financial Receipt Restoration and Reasoning from Degraded Images: Datasets, Methods and Results    }}, author={    Guan, Bochen and  Li, Jinlong and  Yang, Kangning and  Ke, Chuang and  Cai, Jie and  Vasluianu, Florin and  Timofte, Radu and others    },   booktitle={CVPR (CVPR) Workshops},  
year = {2026} 
}

@inproceedings{ntire26faceres, 
title={{    The Second Challenge on Real-World Face Restoration at NTIRE 2026: Methods and Results    }}, 
author={    Wang, Jingkai and  Gong, Jue and  Chen, Zheng and  Liu, Kai and  Li, Jiatong and  Zhang, Yulun and  Timofte, Radu and  others    },
booktitle={CVPR (CVPR) Workshops},  
year = {2026} 
}

@inproceedings{ntire26reflection, 
title={{    NTIRE 2026 Challenge on Single Image Reflection Removal in the Wild: Datasets, Results, and Methods    }}, 
author={    Cai, Jie and  Yang, Kangning and  Li, Zhiyuan and  Vasluianu, Florin and  Timofte, Radu and others    },   
booktitle={CVPR (CVPR) Workshops},  
year = {2026} 
}

@inproceedings{ntire26anomalydet, 
title={{    NTIRE 2026  Challenge Report on Anomaly Detection of Face Enhancement for UGC Images    }}, 
author={    Zhong, Yan and   Ma,  Qiufang and  Wang, Zhen and  Jiang, Tingting and  Timofte, Radu and others    },   
booktitle={CVPR (CVPR) Workshops},  
year = {2026} 
}

@inproceedings{ntire26videosal, 
title={{    NTIRE 2026 Challenge on Video Saliency Prediction: Methods and Results    }}, 
author={    Moskalenko, Andrey and  Bryncev, Alexey and  Kosmynin, Ivan and  Shilovskaya, Kira and  Erofeev, Mikhail and  Vatolin, Dmitry and  Timofte, Radu and others    },   
booktitle={CVPR (CVPR) Workshops},  
year = {2026} 
}

@inproceedings{ntire26effsr, 
title={{The Eleventh NTIRE 2026 Efficient Super-Resolution Challenge Report    }}, 
author={Ren, Bin and  Guo, Hang and  Shu, Yan and  Ma, Jiaqi and  Cui, Ziteng and  Liu, Shuhong  and  Mei, Guofeng  and  Sun, Lei and  Wu, Zongwei and  Khan, Fahad Shahbaz and  Khan, Salman and  Timofte, Radu and  Li, Yawei and others    },   
booktitle={CVPR (CVPR) Workshops},  
year = {2026} 
}

@inproceedings{ntire26realx3d, 
title={{3D Restoration and Reconstruction in Adverse Conditions: RealX3D Challenge Results    }}, 
author={Liu, Shuhong and  Cui, Ziteng and  Bao, Chenyu and  Chu, Xuangeng and  Gu, Lin and  Ren, Bin and  Timofte, Radu and  Conde, Marcos V. and others    },   
booktitle={CVPR (CVPR) Workshops},  
year = {2026} 
}

@inproceedings{ntire26denoising, 
title={{The Third Challenge on Image Denoising at NTIRE 2026: Methods and Results    }}, 
author={Sun, Lei and  Guo, Hang and  Ren, Bin and  Su, Shaolin and  Wang, Xian and  Pani Paudel, Danda and  Van Gool, Luc and  Timofte, Radu and  Li, Yawei and others    },   
booktitle={CVPR (CVPR) Workshops},  
year = {2026} 
}

@article{zhang2025perceive,
  title={Perceive-ir: Learning to perceive degradation better for all-in-one image restoration},
  author={Zhang, Xu and Ma, Jiaqi and Wang, Guoli and Zhang, Qian and Zhang, Huan and Zhang, Lefei},
  journal={TIP},
  year={2025},
  publisher={IEEE}
}

@inproceedings{ntire26aberration, 
title={{    NTIRE 2026 The First Challenge on Blind Computational Aberration Correction: Methods and Results    }}, 
author={    Sun, Lei and  Qian, Xiaolong and  Jiang, Qi and  Wang, Xian and  Gao, Yao and  Yang, Kailun and  Wang, Kaiwei and  Timofte, Radu and  Pani Paudel, Danda and  Van Gool, Luc and others    },   
booktitle={CVPR (CVPR) Workshops},  
year = {2026} 
}

@inproceedings{ntire26eventblurr, 
title={{    The Second Challenge on Event-Based Image Deblurring at NTIRE 2026: Methods and Results    }}, 
author={    Sun, Lei and  Li, Weilun and  Wang, Xian and  Li, Zhendong and  Shi, Letian and  Xu, Dannong and  Zhang, Deheng and  Hu, Mengshun and  Guo, Shuang and  Su, Shaolin and  Timofte, Radu and  Pani Paudel, Danda and  Van Gool, Luc and others    },   
booktitle={CVPR (CVPR) Workshops},  
year = {2026} 
}

@inproceedings{ntire26bursthdr, 
title={{    NTIRE 2026 Challenge on Efficient Burst HDR and Restoration: Datasets, Methods, and Results    }}, 
author={    Park, Hyunhee and  Park, Eunpil and  Lee, Sangmin and  Timofte, Radu and others    },   
booktitle={CVPR (CVPR) Workshops},  
year = {2026} 
}

@inproceedings{ntire26twilight, 
title={{    NTIRE 2026 Low-light Enhancement: Twilight Cowboy Challenge    }}, 
author={    Khalin, Aleksei and  Ershov, Egor and  Panshin, Artem and  Korchagin, Sergey and  Lobarev, Georgiy and  Terekhin, Arseniy and  Dorogova, Sofiia and  Shamsutdinov, Amir and  Mamedov, Yasin and  Khalfin, Bakhtiyar and  Sheludko, Bogdan and  Zilyaev, Emil and  Banić, Nikola and  Perevozchikov, Georgy and  Timofte, Radu and others    },   
booktitle={CVPR (CVPR) Workshops},  
year = {2026} 
}

@inproceedings{ntire26effllie, 
title={{    Efficient Low Light Image Enhancement: NTIRE 2026 Challenge Report    }}, 
author={    Yan, Jiebin  and  Tu, Chenyu  and  Lin, Qinghua and  WU, Zongwei and  Zhang , Weixia and  Wang, Zhihua and  Cao, Peibei and  Fang, Yuming  and  Liu, Xiaoning  and  Zhou, Zhuyun and  Timofte, Radu  and  others    },   
booktitle={CVPR (CVPR) Workshops},  
year = {2026} 
}

@article{sultan2025hybridatnet,
  title={HybridATNet: Multi-Scale Attention and Hybrid Feature Refinement Network for Remote Sensing Image Super-Resolution},
  author={Sultan, Naveed  and  Ruangsang, Watchara  and  Aramvith, Supavadee },
  journal={IEEE Access},
  volume={13},
  year={2025}
}

@inproceedings{shu2026terrascope,
  title={TerraScope: Pixel-Grounded Visual Reasoning for Earth Observation}, 
  author={Yan Shu and Bin Ren and Zhitong Xiong and Xiao Xiang Zhu and Begüm Demir and Nicu Sebe and Paolo Rota},
  year={2026},
  booktitle={CVPR},
}

@inproceedings{li2026chorus,
  title={Chorus: Multi-Teacher Pretraining for Holistic 3D Gaussian Scene Encoding}, 
  author={Yue Li and Qi Ma and Runyi Yang and Mengjiao Ma and Bin Ren and Nikola Popovic and Nicu Sebe and Theo Gevers and Luc Van Gool and Danda Pani Paudel and Martin R Oswald},
  year={2026},
  booktitle={CVPR},
}

@article{tang2025degradation,
  title={Degradation-aware residual-conditioned optimal transport for unified image restoration},
  author={Tang, Xiaole and Gu, Xiang and He, Xiaoyi and Hu, Xin and Sun, Jian},
  journal={TPAMI},
  year={2025},
  publisher={IEEE}
}

@article{xie2025mat,
  title={Mat: Multi-range attention transformer for efficient image super-resolution},
  author={Xie, Chengxing and Zhang, Xiaoming and Li, Linze and Fu, Yuqian and Gong, Biao and Li, Tianrui and Zhang, Kai},
  journal={TCSVT},
  year={2025},
  publisher={IEEE}
}

@inproceedings{sun2023safmn,
    title={Spatially-Adaptive Feature Modulation for Efficient Image Super-Resolution},
    author={Sun, Long and Dong, Jiangxin and Tang, Jinhui and Pan, Jinshan},
    booktitle={ICCV},
    year={2023}
}

@inproceedings{zheng2026open,
  title={Open-World Deepfake Attribution via Confidence-Aware Asymmetric Learning},
  author={Zheng, Haiyang and Pu, Nan and Li, Wenjing and Long, Teng and Sebe, Nicu and Zhong, Zhun},
  booktitle={AAAI},
  volume={40},
  pages={13378--13386},
  year={2026}
}

@inproceedings{vim,
  title={Vision Mamba: Efficient Visual Representation Learning with Bidirectional State Space Model},
  author={Zhu, Lianghui and Liao, Bencheng and Zhang, Qian and Wang, Xinlong and Liu, Wenyu and Wang, Xinggang},
  booktitle={ICML},
  year={2024}
}

@inproceedings{liu2018rethinking,
  title={Rethinking the Value of Network Pruning},
  author={Liu, Zhuang and Sun, Mingjie and Zhou, Tinghui and Huang, Gao and Darrell, Trevor},
  booktitle={ICLR},
  year={2019}
}

@article{lobba2025inverse,
  title={Inverse Virtual Try-On: Generating Multi-Category Product-Style Images from Clothed Individuals},
  author={Lobba, Davide and Sanguigni, Fulvio and Ren, Bin and Cornia, Marcella and Cucchiara, Rita and Sebe, Nicu},
  journal={ICLR},
  year={2026}
}

@article{ren2024sharing,
  title={Sharing key semantics in transformer makes efficient image restoration},
  author={Ren, Bin and Li, Yawei and Liang, Jingyun and Ranjan, Rakesh and Liu, Mengyuan and Cucchiara, Rita and Gool, Luc V and Yang, Ming-Hsuan and Sebe, Nicu},
  journal={NeurIPS},
  volume={37},
  pages={7427--7463},
  year={2024}
}

@article{zheng2025distilling,
  title={Distilling efficient vision transformers from cnns for semantic segmentation},
  author={Zheng, Xu and Luo, Yunhao and Zhou, Pengyuan and Wang, Lin},
  journal={PR},
  volume={158},
  pages={111029},
  year={2025},
  publisher={Elsevier}
}

@inproceedings{zhao2024denoising,
  title={Denoising diffusion probabilistic models for action-conditioned 3d motion generation},
  author={Zhao, Mengyi and Liu, Mengyuan and Ren, Bin and Dai, Shuling and Sebe, Nicu},
  booktitle={ICASSP},
  pages={4225--4229},
  year={2024},
  organization={IEEE}
}

@inproceedings{ma2024shapesplat,
  title={Shapesplat: A large-scale dataset of gaussian splats and their self-supervised pretraining},
  author={Ma, Qi and Li, Yue and Ren, Bin and Sebe, Nicu and Konukoglu, Ender and Gevers, Theo and Van Gool, Luc and Paudel, Danda Pani},
  booktitle={3DV},
  year={2025}
}

@inproceedings{ren2023masked,
  title={Masked jigsaw puzzle: A versatile position embedding for vision transformers},
  author={Ren, Bin and Liu, Yahui and Song, Yue and Bi, Wei and Cucchiara, Rita and Sebe, Nicu and Wang, Wei},
  booktitle={CVPR},
  pages={20382--20391},
  year={2023}
}

@inproceedings{xie2025star,
  title={Star: Spatial-temporal augmentation with text-to-video models for real-world video super-resolution},
  author={Xie, Rui and Liu, Yinhong and Zhou, Penghao and Zhao, Chen and Zhou, Jun and Zhang, Kai and Zhang, Zhenyu and Yang, Jian and Yang, Zhenheng and Tai, Ying},
  booktitle={ICCV},
  pages={17108--17118},
  year={2025}
}

@inproceedings{yu2017compressing,
  title={On compressing deep models by low rank and sparse decomposition},
  author={Yu, Xiyu and Liu, Tongliang and Wang, Xinchao and Tao, Dacheng},
  booktitle={CVPR},
  pages={7370--7379},
  year={2017}
}

@inproceedings{georgescu2023multimodal,
  title={Multimodal multi-head convolutional attention with various kernel sizes for medical image super-resolution},
  author={Georgescu, Mariana-Iuliana and Ionescu, Radu Tudor and Miron, Andreea-Iuliana and Savencu, Olivian and Ristea, Nicolae-C{\u{a}}t{\u{a}}lin and Verga, Nicolae and Khan, Fahad Shahbaz},
  booktitle={CVPR},
  pages={2195--2205},
  year={2023}
}

@article{xiao2024frequency,
  title={Frequency-assisted mamba for remote sensing image super-resolution},
  author={Xiao, Yi and Yuan, Qiangqiang and Jiang, Kui and Chen, Yuzeng and Zhang, Qiang and Lin, Chia-Wen},
  journal={TMM},
  volume={27},
  pages={1783--1796},
  year={2024},
  publisher={IEEE}
}
